\documentclass[preprint,review,12pt]{elsarticle}

\usepackage{amssymb}
\usepackage{amsmath,amsfonts}
\usepackage{algorithmic}
\usepackage{algorithm}
\usepackage{array}
\usepackage[caption=false,font=normalsize,labelfont=sf,textfont=sf]{subfig}
\usepackage{textcomp}
\usepackage{stfloats}
\usepackage{url}
\usepackage{verbatim}
\usepackage{graphicx}
\usepackage{color}

\usepackage{balance}
\usepackage{multirow}
\usepackage{booktabs}

\newcommand{\revise}[1]{\textcolor{black}{#1}}

\journal{PATTERN RECOGNITION}

\begin{document}

\begin{frontmatter}



\title{ARAI-MVSNet: A multi-view stereo depth estimation network with adaptive depth range and depth interval}

\author[1,2,3]{Song Zhang}

\author[4]{Wenjia Xu}

\author[1,2]{Zhiwei Wei\corref{mycorrespondingauthor}}


\author[1,2]{Lili Zhang}
\author[1,2]{Yang Wang}
\author[1,2]{Junyi Liu}

\cortext[mycorrespondingauthor]{Corresponding authors}

\affiliation[1]{Aerospace Information Research Institute, Chinese Academy of Sciences, Beijing 100190, China}

\affiliation[2]{Key Laboratory of Network Information System Technology(NIST), Institute of Electronics, Chinese Academy of Sciences, Beijing 100190, China}

\affiliation[3]{School of Electronic, Electrical and Communication Engineering, University of Chinese Academy of Sciences, Beijing 100190, China}

\affiliation[4]{State Key Laboratory of Networking and Switching Technology, Beijing University of Posts and Telecommunications, Beijing 100876, China}

\begin{abstract}
    Multi-View Stereo~(MVS) is a fundamental problem in geometric computer vision which aims to reconstruct a scene using multi-view images with known camera parameters. However, the mainstream approaches represent the scene with a fixed all-pixel depth range and equal depth interval partition, which will result in inadequate utilization of depth planes and imprecise depth estimation. In this paper, we present a novel multi-stage coarse-to-fine framework to achieve adaptive all-pixel depth range and depth interval. We predict a coarse depth map in the first stage, then an Adaptive Depth Range Prediction module is proposed in the second stage to zoom in the scene by leveraging the reference image and the obtained depth map in the first stage and predict a more accurate all-pixel depth range for the following stages. In the third and fourth stages, we propose an Adaptive Depth Interval Adjustment module to achieve adaptive variable interval partition for pixel-wise depth range. The depth interval distribution in this module is normalized by Z-score, which can allocate dense depth hypothesis planes around the potential ground truth depth value and vice versa to achieve more accurate depth estimation. Extensive experiments on four widely used benchmark datasets~(DTU, TnT, BlendedMVS, ETH 3D) demonstrate that our model achieves state-of-the-art performance and yields competitive generalization ability. Particularly, our method achieves the highest Acc and Overall on the DTU dataset, while attaining the highest Recall and $F_{1}$-score on the Tanks and Temples intermediate and advanced dataset. Moreover, our method also achieves the lowest $e_{1}$ and $e_{3}$ on the BlendedMVS dataset and the highest Acc and $F_{1}$-score on the ETH 3D dataset, surpassing all listed methods.
\end{abstract}




\end{frontmatter}


\section{Introduction}

\begin{figure}[!ht]
\centering
\includegraphics[width=0.55\textwidth]{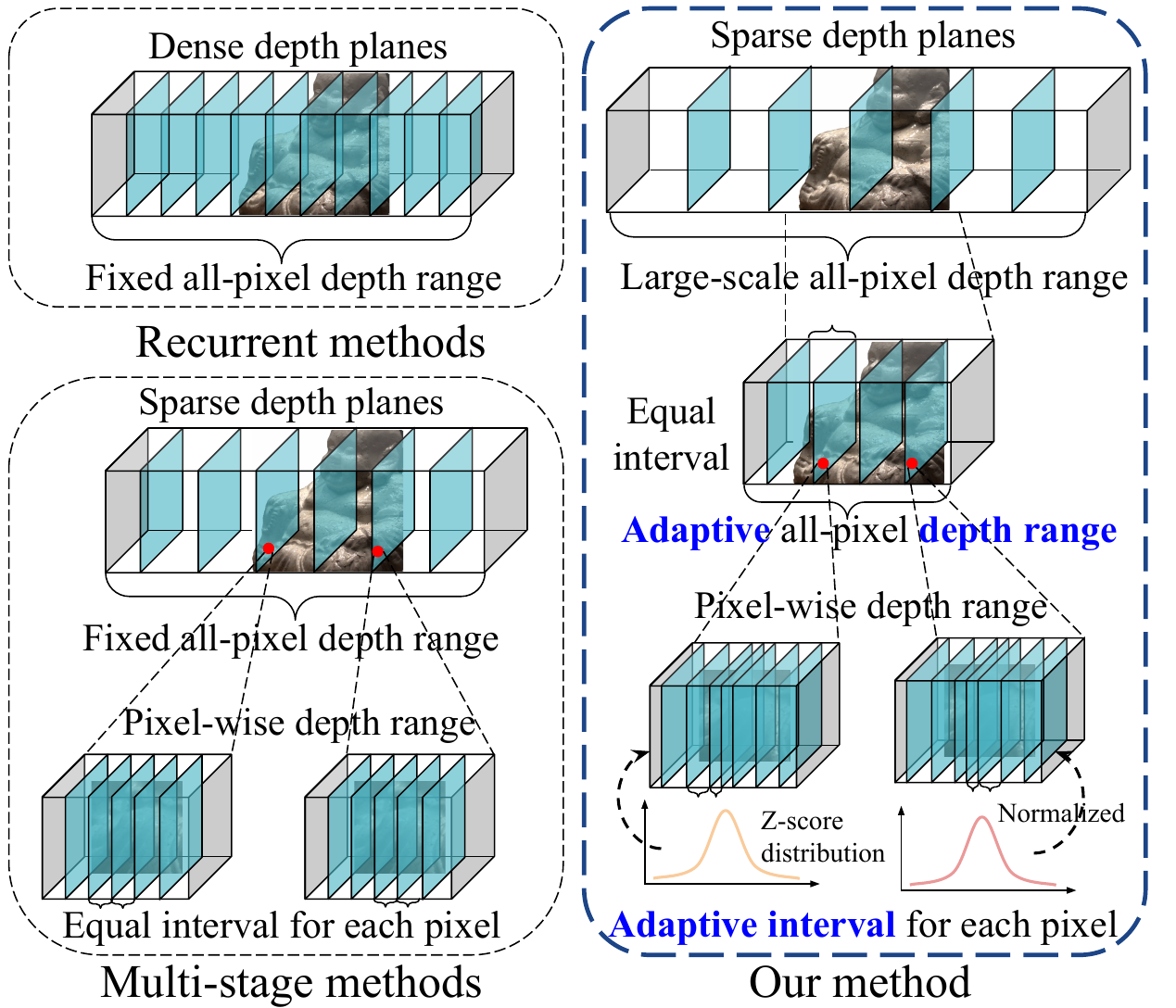}
\caption{Comparisons of the recurrent methods, multi-stage methods and our proposed method. We adopt an all-pixel adaptive depth range and adaptive variable interval partition for pixel-wise depth range.}
\label{fig:Figure 0}
\end{figure}

Multi-View Stereo~(MVS) is a fundamental problem in computer vision that aims to reconstruct a 3D scene from a collection of multiple images captured from different viewpoints. It addresses the challenge of inferring the 3D geometry of a scene by leveraging the visual information contained in these images. The MVS problem has garnered significant attention due to its wide range of applications, including 3D reconstruction \cite{b64}, robot navigation \cite{b65, b67} and so on. To tackle the MVS problem, various techniques have been developed over the years. Early traditional MVS methods \cite{b21} leveraging multi-view consistency measures have achieved considerable performance, but they still demonstrate limited representation capability in low-texture regions, reflections, etc \cite{b1}. Recent advancements in deep learning have revolutionized the field of MVS. MVSNet \cite{b1} that utilizes 2D CNN to extract features and uses 3D CNN to regularize the cost volume yields significant improvements on many MVS benchmarks. However, the pipeline still suffers from low efficiency and high memory consumption. Following works which aim to balance the effectiveness and the efficiency are mainly categorized as recurrent methods \cite{b2,b38} and multi-stage methods \cite{b4,b6}. The recurrent methods utilize the GRU or convolutional LSTM to regularize the cost volume sequentially. While the multi-stage methods propose a novel cascade formulation using multi-scale cost volumes. Following a similar cascade strategy, we have designed our framework to achieve an effective and efficient MVSNet.

The basic idea of our framework is that the surfaces of the objects are usually heterogeneous and their sizes are diverse. However, the above methods all represent the object with a fixed all-pixel depth range and equal depth interval partition and will result in two problems. First, since the objects may have different sizes, the fixed all-pixel depth range may not be a suitable fit for the range of each object, as depicted in Fig.\ref{fig:Figure 0}. As a result, if the fixed all-pixel depth range is larger than the actual depth range of the object, certain depth hypothesis planes may become redundant~(the higher left part of Fig.\ref{fig:Figure 0}), whereas objects that fall beyond the fixed depth range cannot be reconstructed and may lead to a subpar reconstruction quality. Second, since the object surface is usually heterogeneous, employing the equal depth interval partition strategy can result in disparities between predicted depth values and the actual ground truth (GT) depth values~(the lower left part of Fig.\ref{fig:Figure 0}). Insufficient depth hypothesis planes in close proximity to the GT depth value can hinder the ability to accurately estimate depth values, ultimately leading to the imprecise reconstruction of specific details and negatively impacting the overall quality of the reconstruction.

To tackle the aforementioned problems, we propose a novel multi-stage coarse-to-fine framework named ARAI-MVSNet, which mainly consists of two novel modules. As depicted in Fig.\ref{fig:Figure 0}, instead of using a common fixed all-pixel depth range, we propose an innovative Adaptive Depth Range Prediction~(ADRP) module, which enables a more precise zoom-in of the scene. The ADRP module predicts a more accurate all-pixel depth range for each object from a large-scale depth range by leveraging the reference image and the depth map of the former stage. This mechanism effectively utilizes the depth hypothesis planes, leading to the attainment of superior reconstruction outcomes, as evidenced by its high quality. In line with previous studies such as UCS-Net \cite{b11} and CFNet \cite{b54}, we establish the pixel-wise depth range from the all-pixel depth range. However, in contrast to the equal depth interval partition employed in UCS-Net and CFNet, we introduce an Adaptive Depth Interval Adjustment (ADIA) module to reallocate more depth hypothesis planes close to the potential GT depth value. This mechanism utilizes the Z-score formulation \cite{b9} to calculate the offset for each depth plane, taking advantage of the depth map from the previous stage to achieve adaptive depth interval partition for the pixel-wise depth range. Additionally, it is also non-parametric to achieve adaptive depth interval for pixel-wise depth range, thus can significantly improve the efficiency of our module with performance maintained. Furthermore, to extract robust image features, we also design an Atrous Spatial Pyramid Feature Extraction Network~(ASPFNet), where the context-aware features are extracted and fused with larger receptive fields.

We evaluate our method on four different competitive benchmarks to demonstrate the SOTA performance and generalization ability. The results demonstrate that our model achieves the state-of-the-art Acc score and Overall score compared to other pioneer works on DTU dataset, the highest Recall score and $F_{1}$-score score on Tanks and Temples intermediate and advanced dataset respectively. Moreover, our method also achieve the lowest $e_{1}$ score and $e_{3}$ score on the BlendedMVS dataset and the highest Acc score and $F_{1}$-score on the ETH 3D dataset, surpassing all listed methods.

In summary, our main contributions are four folds: (1)~We present a novel multi-stage coarse-to-fine framework to achieve more accurate depth estimation. (2)~We propose an ADRP module to predict an adaptive all-pixel depth range for more reliable reconstruction. (3)~We propose an ADIA module to achieve adaptive variable depth interval partition to estimate more accurate depth values. (4)~We evaluate our method on four benchmarks and significantly advance the state-of-the-art methods in evaluation metrics.

\section{Related Work}
\subsection{Traditional MVS methods}
The core idea of MVS is to estimate the depth or disparity map for each pixel in the images, which represents the corresponding 3D point's distance from the camera(s). This estimation is achieved by analyzing the spatial and photometric consistencies across the multiple views \cite{b69}. The key assumption is that the same 3D point in the scene should project to similar positions in different images, and its appearance should exhibit consistent color or intensity values. By exploiting these correspondences, the MVS algorithms can triangulate the 3D positions of the scene points and reconstruct a dense representation of the underlying geometry \cite{b69}. 

The taxonomic classification of these traditional MVS algorithms can be divided into four main categories: volumetric-based \cite{b12}, mesh-based \cite{b16}, point cloud-based \cite{b15}, and depth map-based \cite{b20}. Volumetric, mesh, and point cloud representations directly visualize the scene, while depth map-based methods estimate depth maps of each reference image and fuse them into other representations. This decouples the complex 3D reconstruction problem into a 2D depth map estimation problem, making depth map-based methods more flexible and robust in MVS \cite{b25,b26}. Despite the fact that traditional MVS methods can obtain impressive results, handcrafted-features based MVS methods still face significant challenges for regions with weak textures and non-Lambertian surfaces \cite{b39}. Therefore, our method mainly focuses on further improving the performance of depth estimation by developing a deep learning-based MVS pipeline.

\subsection{Deep learning-based MVS methods}
Recently, to overcome the blemish of traditional MVS methods, many deep learning-based methods \cite{b1, b62, b61} have been proposed and achieve impressive performances. MVSNet \cite{b1} implements a differentiable homography warping to transform the traditional MVS domain into the deep-learning based domain. This approach involves matching the feature points between different views to achieve spatial and photometric consistencies (it constructs a cost volume) across multiple views. Subsequently, 3D CNNs are employed to regularize the cost volume. However, the 3D CNNs greatly increases memory consumption. R-MVSNet \cite{b2} and $D^{2}$HC-RMVSNet \cite{b38} leverage convolutional GRUs and LSTMs to sequentially regularize the cost volume to avoid using memory-intensive 3D CNNs. However, these methods require a large number of depth hypothesis planes to obtain better performance. And it is difficult to construct a high precision cost volume based on the high-resolution image limited by the memory bottleneck. Therefore, multi-stage methods \cite{b4, b6} have been proposed to obtain high-quality depth maps without increasing memory. 
UCS-Net \cite{b11} uses adaptive thin volumes based on uncertainties of pixel-wise depth predictions. ADR-MVSNet \cite{b62} introduces an Adaptive Depth Reduction module that utilizes the probability distribution of pixels in the depth direction to dynamically learn distinct depth ranges for each pixel. Prior-Net \cite{b61} uses a three-stage cascade strategy in its fast version to efficiently compute a low-resolution depth map. Additionally, a Refine-Net is proposed to upsample the depth and confidence maps to the original high-resolution, further enhancing computational efficiency. Nevertheless, learning-based MVS methods still face challenging issues, such as resource redundancy and limited capability in low-texture regions.

\section{Methodology}
\subsection{Problem Formulation}
\label{sec:PF}
In the deep learning-based MVS task, the goal is to use an end-to-end trainable model to infer a depth map $\mathbf{L}$ from $ N-1 $ adjacent views with their corresponding camera poses. Assume the reference image $\mathbf{I}_{1}$ and source images $\left\{ \mathbf{I}_{i} \right\}_{i=2}^{N}$, where the features $\left\{ \mathbf{F}_{i} \right\}_{i=1}^{N}$ are extracted from them. Then the differentiable homography warping based on the depth hypothesis planes $ \mathbf{D} $ and the extracted $\left\{ \mathbf{F}_{i} \right\}_{i=1}^{N}$ is used to construct a cost volume $ \mathbf{V} $. After regularization process, the regularized cost volume $\mathbf{V}_{reg}$ regresses a probability volume $\mathbf{P}$ by softmax operation. Finally, the global depth map $\mathbf{L}$ is calculated by the probability volume $\mathbf{P}$ and depth hypothesis planes $ \mathbf{D} = [\mathbf{d}_{min},...,\mathbf{d}_{max}] $. And $\mathbf{d}$ represents a depth value among the depth hypothesis planes.

\begin{figure*}[!t]
\centering
\includegraphics[width=0.95\textwidth]{ 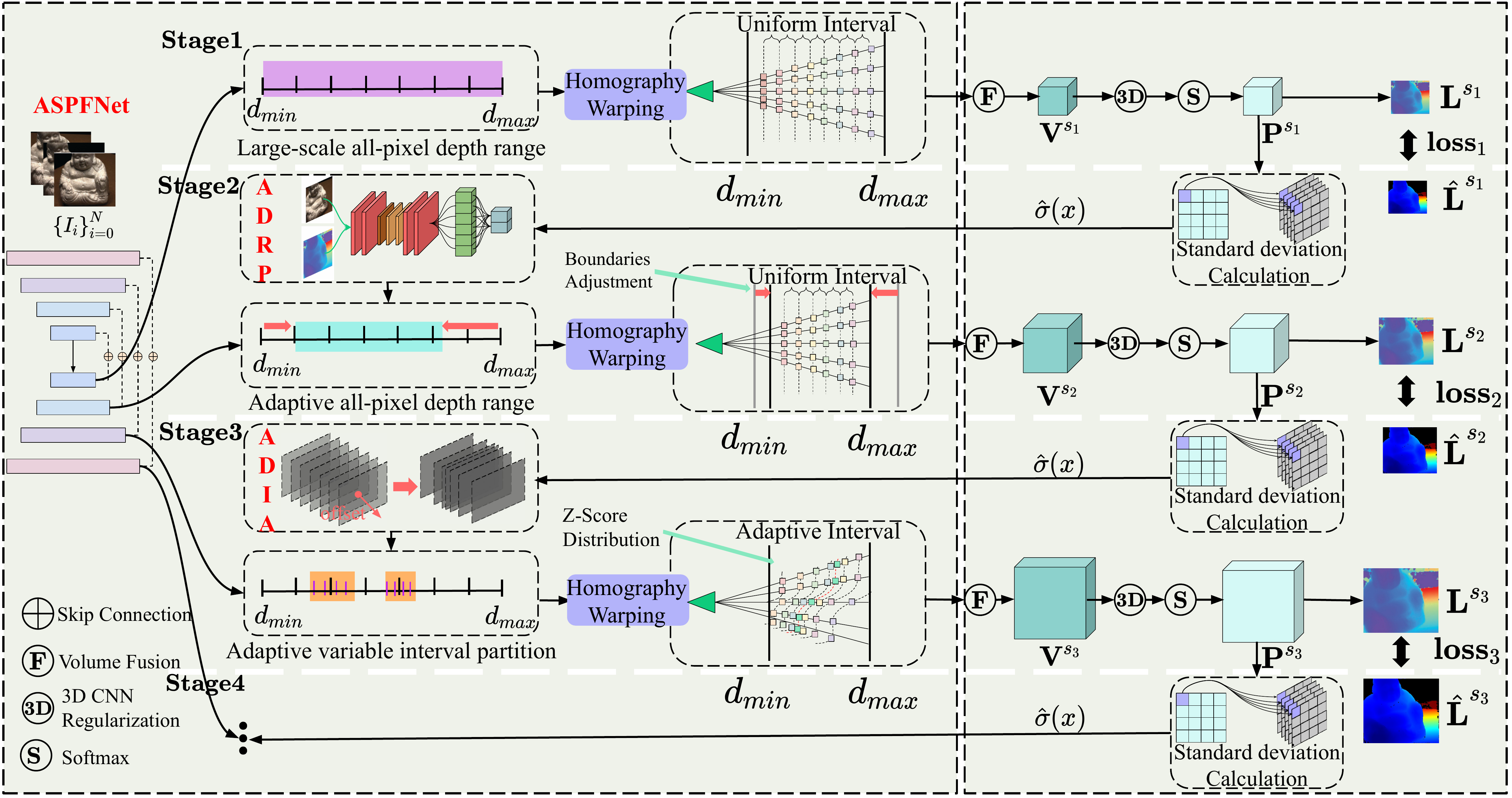}
\caption{Illustration of overall ARAI-MVSNet. This is a typical multi-stage coarse-to-fine framework. The left part is our novel modules, and the right part is inherited from existing methods. The three dots represent the stage 4 is a repeat of the stage 3 process.}
\label{fig:Figure 1}
\end{figure*}

\subsection{ARAI-MVSNet}
\label{sec:NO}

The overall framework of ARAI-MVSNet is shown in Fig.\ref{fig:Figure 1}. It employs a four-stage coarse-to-fine approach for depth map inference. Prior to entering the pipeline, a feature extraction network called ASPFNet is used to extract multi-scale context features. \textbf{Stage 1:} we use the main part of MVSNet based on a large-scale all-pixel depth range to obtain a coarse depth map. \textbf{Stage 2:} we propose an Adaptive Depth Range Prediction~(ADRP) module by leveraging the reference image and the obtained coarse depth map in stage 1 to predict an adaptive all-pixel depth range, then the standard flow is executed as above based on the new all-pixel depth range.
\textbf{Stage 3:} we propose an Adaptive Depth Interval Adjustment~(ADIA) module to adaptively adjust the depth interval for pixel-wise depth range based on Z-score distribution. Then the standard flow is executed as above based on the new depth interval partition.
\textbf{Stage 4:} Stage 4 is similar to Stage 3, but with a greater number of depth hypothesis planes for improved performance.

\subsection{Atrous Spatial Pyramid Feature Extraction Network~(ASPFNet)}
\label{sec:ASPF}

Conventional CNNs face challenges in effectively handling reflective surfaces, low-textured regions, and texture-less regions when operating on standard 2D grids with fixed receptive fields \cite{b39}. Therefore, to address this challenge and extract more comprehensive image features, we have designed an Atrous Spatial Pyramid Feature Extraction Network (ASPFNet) that combines dilated convolution and FPN to enlarge the receptive field and obtain better multi-scale features. The ASPFNet is an encoder-decoder architecture~(as shown in Fig.\ref{fig:Figure 1}). The ASPFNet is composed of four Encoder Downsample Blocks~(EDB), four Decoder UpSample Blocks~(DUB) and skip connections. EDB shrinks the feature map and enlarges the receptive field with dilated convolution. DUB enlarges the feature map and fuses high-dimensional features from skip connections. This bottleneck encoder-decoder architecture can integrate multi-level features by simultaneously fusing low- and high-level information to achieve high-quality reconstruction. More details of ASPFNet are in the supplementary material.

\subsection{Adaptive Depth Range Prediction~(ADRP)}
To achieve the image-level uniform adjustment, we utilize a large-scale all-pixel depth range to obtain a coarse depth map in stage 1. Since the large-scale all-pixel depth map may not be accurate enough to support subsequent stages, we propose an Adaptive Depth Range Prediction~(ADRP) module in stage 2 that calculates an adaptive all-pixel depth range by leveraging the information from the coarse depth map of stage 1 and the reference image. 

\label{sec:ADRP}
\begin{figure}[!t]
\centering
\includegraphics[width=0.7\textwidth]{ 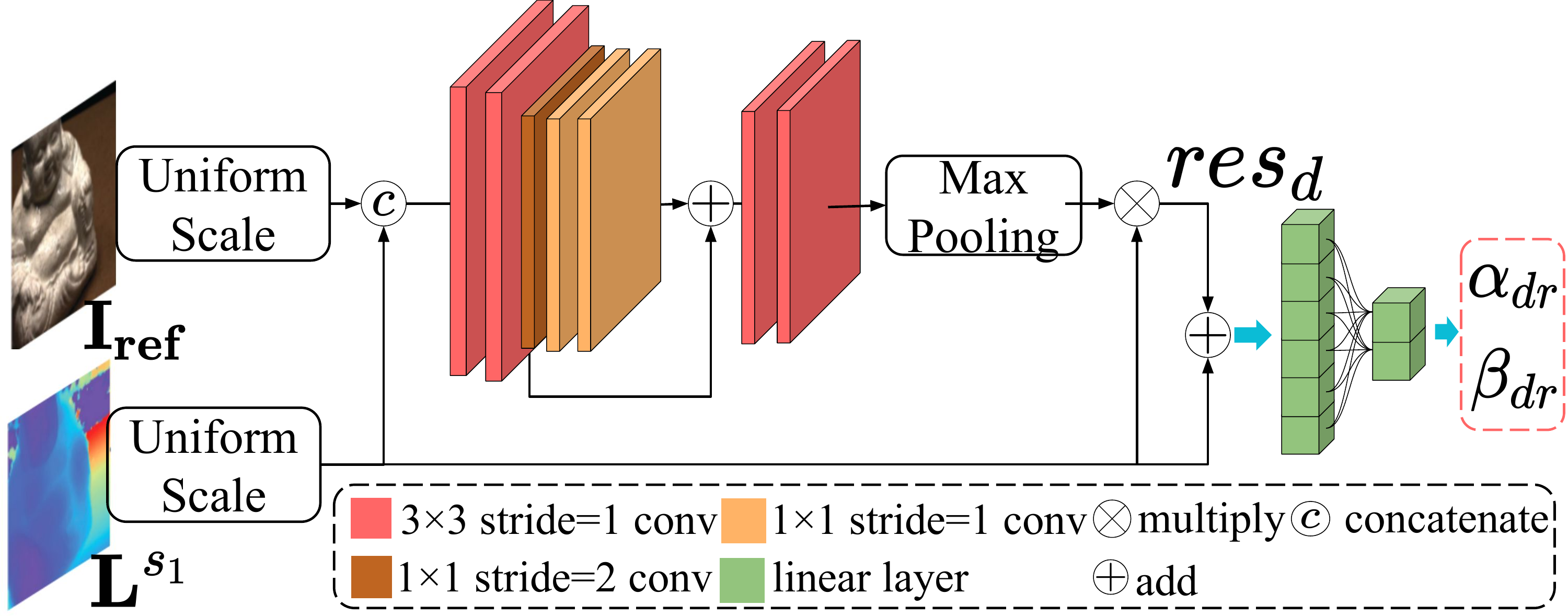}
\caption{Illustration of Scalar Calculation Network.}
\label{fig:Figure 3}
\end{figure}

In stage 1, we establish a large-scale all-pixel depth range to derive a preliminary coarse depth map denoted as $\mathbf{L}^{s_{1}}$. The maximum and minimum depth values of $\mathbf{L}^{s_{1}}$, represented by $\mathbf{L}^{s_{1}}(x_{min})$ and $\mathbf{L}^{s_{1}}(x_{max})$, respectively, are then utilized as the initial depth boundaries in stage 2. Where the $x$ represents a pixel in the depth map, $x_{max}$ and $x_{min}$ denote the position of the pixel of the maximum and minimum depth values in the depth map, respectively.
Using the reference image $\mathbf{I}_{ref}$ and coarse depth map $\mathbf{L}^{s{1}}$ as inputs, we predict an updated and more precise all-pixel depth range by designing a Scalar Calculation Network~(SCN)~(illustrated in Figure 3) to calculate $ \alpha_{dr} $ and $ \beta_{dr} $ for adjusting the depth boundaries:
\begin{equation}
    \label{eq:equation_2}
    \begin{aligned}
        & \mathbf{d}^{s_{2}}_{min} = \mathbf{L}^{s_{1}}(x_{min}) + \alpha_{dr} \times \hat{\mathbf{\sigma}}(x_{min}) \\
        & \mathbf{d}^{s_{2}}_{max} = \mathbf{L}^{s_{1}}(x_{max}) + \beta_{dr} \times \hat{\mathbf{\sigma}}(x_{max}) \, \\
    \end{aligned}
\end{equation}
where $\mathbf{\hat{\sigma}}(x_{min})$ and $\mathbf{\hat{\sigma}}(x_{max})$, working as the adjustment factors, are the standard deviations corresponding to positions. As the loss decreases, our depth estimation becomes more accurate, which reduces $\mathbf{\hat{\sigma}}(x_{min})$ over time. Thus, by adjusting the coarse depth range while maintaining the loss constraint, we can predict a more precise adaptive all-pixel depth range.

In particular, we consider the depth value as the mean value of previous stage, and $\mathbf{\hat{\sigma}}(x_{min})$ and $\mathbf{\hat{\sigma}}(x_{max})$ are calculated as Eq. \eqref{eq:equation_1}.
\begin{equation}
    \label{eq:equation_1}
    \hat{\mathbf{\sigma}}(x_{i})=\sqrt{\sum_{j}^{D} \mathbf{P}^{s_{1}}_{j}(x_{i}) \cdot\left(\mathbf{d}^{s_{1}}_{j}(x_{i})-\mathbf{L}^{s_{1}}(x_{i})\right)^{2}} \,
\end{equation}
with $x_{i} \in [x_{min}, x_{max}]$, where $ \mathbf{P}^{s_{1}}_{j} $ expresses pixel-wise depth probability distributions of stage 1, $ \mathbf{d}^{s_{1}}_{j} $ represents the depth hypotheses plane of stage 1, and $ \mathbf{P}^{s_{1}}_{j}(x_{i}) $ represents how probable the depth at pixel $ x_{i} $ is $ \mathbf{d}^{s_{1}}_{j}(x_{i}) $. 

We concatenate the coarse depth map and reference image, and leverage stacked convolutional layers $ \textbf{w}_{1} $
to extract depth residual information $ res_{d} $. We add $ res_{d} $ to the depth map, and regress the final scalars $ \alpha_{dr}, \beta_{dr} $ with a linear layer $ \textbf{w}_{2} $:

\begin{equation}
    \label{eq:equation_1_1}
    \begin{aligned}
        & res_{d} = maxp(\textbf{w}_{1}([\mathbf{L^{s1}}\,, \mathbf{I_{ref}}])\odot \mathbf{L}^{s1}  \,, \\
        & \alpha_{dr}, \beta_{dr} = \textbf{w}_{2}(res_{d} + \mathbf{L}^{s1}) \,,
    \end{aligned}
\end{equation}

After obtaining the adaptive all-pixel depth range, we partition the all-pixel depth range by equal interval partition to obtain depth hypothesis planes $ \mathbf{D}^{s2} = [\mathbf{d}^{s_{2}}_{min},...,\mathbf{d}^{s_{2}}_{max}] $ to construct the cost volume in stage 2. Based on the cost volume, then a cost volume regularization network (similar to the 3D UNet used in MVSNet \cite{b1}) is employed to regress a depth map $\mathbf{L}^{s2}$ in stage 2.

\subsection{Adaptive Depth Interval Adjustment~(ADIA)}
\label{sec:ADIA}
\begin{figure}[!t]
\centering
\includegraphics[width=0.7\textwidth]{ 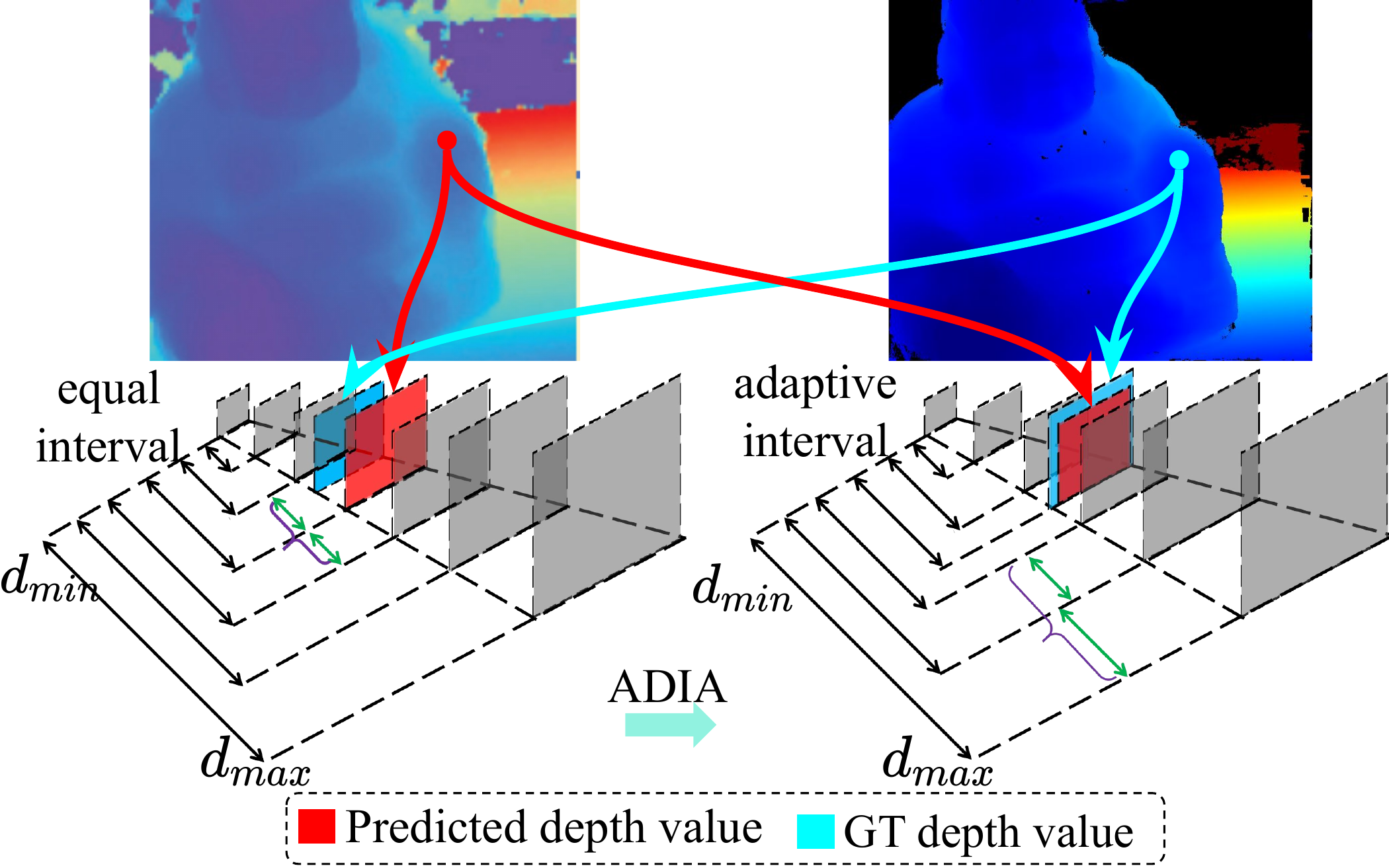}
\caption{The scheme of Adaptive Depth Interval Adjustment Module.}
\label{fig:Figure 4}
\end{figure}

Instead of employing equal interval partition for pixel-wise depth estimation, we propose an adaptive variable interval partition strategy to allocate more dense depth hypothesis planes near the potential ground truth depth value, and allocate relatively sparse depth hypothesis planes away from it. Z-score is a measure of the relative position, which can measure the distance from the values to the mean (namely the GT depth value).
Thus, we follow the Z-score distribution \cite{b9} to allocate the position of depth hypothesis planes to achieve the above strategy. To accomplish this, we propose an Adaptive Depth Interval Adjustment (ADIA) module (as illustrated in Fig. \ref{fig:Figure 4}). The computation procedure is delineated in Eq.\eqref{eq:equation_8}, in which we calculate different weights for equal intervals to obtain variable intervals and add them to the corresponding basic depth hypothesis planes to accomplish adaptive depth interval adjustment.

\begin{equation}
    \label{eq:equation_8}
    \mathbf{d}^{s_{3}}_{i}(x) = \mathbf{d}^{s_{3}}_{i}(x) + \hat{\mathbf{d}}^{s_{3}}_{inter}(x) \times \mathbf{offset}_{i}(x)
\end{equation}

\noindent where $ \mathbf{d}^{s_{3}}_{i}(x) $ represents $i^{th}$ plane of depth hypothesis planes of pixel $x$, $ \hat{\mathbf{d}}^{s_{3}}_{inter}(x) $ is the pixel-wise equal interval, $ \mathbf{offset}_{i}(x) $ is the weight for the equal interval. In Eq. \eqref{eq:equation_8}, we allocate different offsets to depth intervals to achieve interval scaling based on the Z-score \cite{b9} formulation.

The calculation process of $ \hat{\mathbf{d}}^{s_{3}}_{inter}(x) $ and $ \mathbf{offset}_{i}(x) $ is outlined as follows. Specifically, we first utilize previous stage depth map $ \mathbf{L}^{s2} $ and probability volume $ \mathbf{P}^{s2} $ to calculate the pixel-wise standard deviation $\hat{\mathbf{\sigma}}(x)$, which is defined as Eq. \eqref{eq:equation_4}.

\begin{equation}
    \label{eq:equation_4}
    \hat{\mathbf{\sigma}}(x)=\sqrt{\sum_{j}^{D} \mathbf{P}^{s_{2}}_{j}(x) \cdot\left(\mathbf{d}^{s_{2}}_{j}(x)-\mathbf{L}^{s_{2}}(x)\right)^{2}}
\end{equation}

Then we leverage above results to calculate the pixel-wise depth range for current stage, the upper and lower boundaries of the pixel-wise depth range, i.e., pixel-wise depth min $ \mathbf{d}^{s3}_{min}(x) $ and pixel-wise depth max $ \mathbf{d}^{s3}_{max}(x) $ are defined as Eq. \eqref{eq:equation_5}.

\begin{equation}
    \centering
    \label{eq:equation_5}
    \begin{aligned}
        &\mathbf{d}^{s_{3}}_{min}(x) = \mathbf{L}^{s_{2}}(x) - \hat{\mathbf{\sigma}}(x), 
        \mathbf{d}^{s_{3}}_{max}(x) = \mathbf{L}^{s_{2}}(x) + \hat{\mathbf{\sigma}}(x) \\
        &\mathbf{D}^{s3}(x) = [\mathbf{d}^{s_{3}}_{min}(x), \cdot \cdot \cdot , \mathbf{d}^{s_{3}}_{i}(x), \cdot \cdot \cdot, \mathbf{d}^{s_{3}}_{max}(x)]
    \end{aligned}
\end{equation}

\noindent where $ \mathbf{D}^{s3}(x) $ represents the pixel-wise depth hypothesis planes of stage 3. And we partition above pixel-wise depth range to obtain the pixel-wise equal interval $ \hat{\mathbf{d}}^{s_{3}}_{inter}(x) $ as our fixed step. The standard equal interval partition is defined as Eq. \eqref{eq:equation_6}.

\begin{equation}
    \label{eq:equation_6}
    \hat{\mathbf{d}}^{s_{3}}_{inter}(x) = \frac{\mathbf{d}^{s_{3}}_{max}(x) - \mathbf{d}^{s_{3}}_{min}(x)}{\mathbf{D}_{num}}
\end{equation}

\noindent where $ \mathbf{D}_{num} $ represents the number of depth hypothesis planes of stage 3. Inspired by Z-score\cite{b9}, we utilize the depth value of the previous stage~(considered as mean value) and standard deviation to calculate the offset for each plane, and use softmax to achieve normalization. The calculation process is formulated by Eq. \eqref{eq:equation_7}.

\begin{equation}
    \label{eq:equation_7}
    \mathbf{offset}_{i}(x) = softmax(\frac{\mathbf{d}^{s_{3}}_{i}(x) - \mathbf{L}^{s_{2}}(x)}{\hat{\mathbf{\sigma}}(x)}) 
\end{equation}

\noindent where $\mathbf{d}^{s_{3}}_{i}(x)$ represents the $i^{th}$ depth value correspondent to pixel $x$ among the depth hypothesis planes of stage 3, and $\mathbf{L}^{s_{2}}(x)$ represents the depth value of pixel $x$ in predicted depth map of previous stage.
Finally, we use the offset to achieve adaptive variable interval partition, as defined by Eq. \eqref{eq:equation_8}.

\subsection{Training Loss}Following the previous methods \cite{b1}, we adopt same mean absolute difference loss as our loss:

\begin{equation}
    \label{eq:equation_13}
    \begin{aligned}
    & \text { Loss }=\sum_{i=1}^{4} \lambda_{i} \sum_{x \in \mathbf{x}_{\text {valid }}} \|\hat{\mathbf{L}}^{s_{i}}(x)-\mathbf{L}^{s_{i}}(x)\|_{1}, \\\ 
    \end{aligned}
\end{equation}

\noindent where $ \hat{\mathbf{L}}^{s_{i}} $ denotes the GT depth map of each stage, $ \mathbf{L}^{s_{i}} $ denotes the predicted depth map of each stage. We set $\lambda_{i}$ to be 0.5, 1.0, 1.5, 2.0 for each stage.

\section{Experiments}

\subsection{Dataset}
We evaluate our method on four datasets widely used in MVS. These include: (1)~\textbf{DTU} \cite{b28}, captured in a laboratory setting, it consists of 124  scenes and 7 lighting conditions. (2)~\textbf{Tanks and Temples} \cite{b32}, captured from real outdoor sensors, with more complex and realistic scenes; (3)~\textbf{BlendedMVS} \cite{b33}, with over 17k indoor and outdoor images of 113 scenes split into training and testing sets; and (4)~\textbf{ETH 3D} \cite{b34}, with high-resolution calibrated images of scenes containing significant viewpoint variations.

\subsection{Implement Details}
\noindent \textbf{Training}: The proposed ARAI-MVSNet was trained on DTU dataset and finetuned on BlendedMVS dataset. (1)~During training, input image resolution was set to $640 \times 512$ and $N=3$ training views were used. Hypothetical depth planes of four stages were set to $[16,48,16,8]$. The model was optimized for 16 epochs using RMSProp optimizer with a learning rate of 0.0006 and decay weight of 0.001. The batch size was 16 and 8 NVIDIA GTX 2080ti GPUs were used for training. Standard metrics provided by the official evaluation protocol \cite{b28} were used for performance evaluation.
(2)~For finetuning, the best result of DTU was used for 10 epochs with $N=7$ images of original size of $768 \times 576$. The learning rate was set to 0.0005 and the hypothetical depth plane was set to $[16,64,32,8]$.
\noindent \textbf{Testing}:
(1)~For testing on DTU dataset, our best result was used with $N=7$ adjacent images, image resolution of $960 \times 1280$, and hypothetical depth planes of $[16, 64, 16, 8]$.
(2)~To verify the generalization ability on Tanks and Temples benchmark, the finetuned model was used with input image sizes of $1024 \times 1920$ or $1024 \times 2048$, $N=11$ input images, and hypothetical depth planes of $[16, 128, 32, 16]$.

\begin{figure}[!t]
\centering
\includegraphics[width=0.9\textwidth]{ 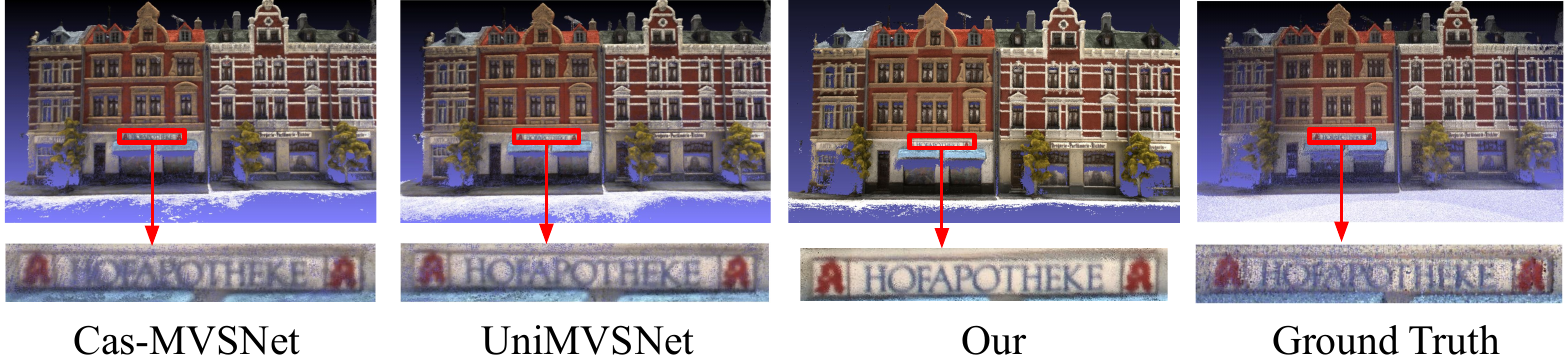}
\caption{Quantitative results of scan 15 on DTU. The top row exhibits the point clouds obtained by different methods. The bottom row exhibits the detailed comparison.}
\label{fig:Figure 5}
\end{figure}

\subsection{Performance on close-range experimental scene benchmark} 
Close-range scenarios are characterized by a short distance between the camera and the scene being captured. Given that the DTU dataset contains small scenes and close camera-to-scene distances, we leveraged this dataset to evaluate the effectiveness of our model on a close-range experimental scene benchmark. Table.\ref{tab:Table.1} shows the quantitative results on DTU dataset. 

\begin{table}[!t]
    \centering
    \LARGE
    \resizebox{8.5cm}{!}{
    \begin{tabular}{c|ccc}
    \toprule
        Methods & Acc. (mm) $\downarrow$ & Comp. (mm) $\downarrow$ & Overall (mm) $\downarrow$ \\ \hline
        MVSNet \cite{b1} & 0.396 & 0.527 & 0.462 \\ 
        R-MVSNet \cite{b2} & 0.383 & 0.452 & 0.417 \\  
        CasMVSNet \cite{b4} & 0.325 & 0.385 & 0.355 \\ 
        CVP-MVSNet \cite{b6} & 0.296 & 0.406 & 0.351 \\ 
        UCS-Net \cite{b11} & 0.338 & 0.349 & 0.344 \\ 
        $D^{2}$HC-RMVSNet \cite{b38} & 0.395 & 0.378 & 0.386 \\ 
        AA-RMVSNet \cite{b39} & 0.376 & 0.339 & 0.357 \\ 
        PatchmatchNet \cite{b49} & 0.427 & 0.277 & 0.352 \\ 
        EPP-MVSNet \cite{b50} & 0.413 & 0.296 & 0.355 \\ 
        ADR-MVSNet \cite{b62} & 0.354 & 0.317 & 0.335 \\
        IterMVS \cite{b51} & 0.373 & 0.354 & 0.363 \\ 
        UniMVSNet \cite{b42} & 0.352 & 0.278 & 0.315 \\ 
        NP-CVP-MVSNet \cite{b52} & 0.356 & \textbf{0.275} & 0.315 \\ 
        Prior-Net \cite{b61} & 0.351 & 0.287 & 0.319 \\
        \midrule
        Our & \textbf{0.292} & 0.334 & \textbf{0.313} \\
    \bottomrule
    \end{tabular}
    }
    \caption{Quantitative results of different methods on DTU's evaluation set. } 
    \label{tab:Table.1}
\end{table}

From Table.\ref{tab:Table.1}, we have the following observations: our ARAI-MVSNet achieves the SOTA Acc score. Specifically, our ARAI-MVSNet improves the Acc from 0.325 of CasMVSNet \cite{b4} to 0.292, 0.354 of ADR-MVSNet \cite{b62} to 0.292, 0.351 of Prior-Net \cite{b61} to 0.292 and 0.352 of UniMVSNet \cite{b42} to 0.292~($ 1^{st} $), meanwhile higher than previous deep learning-based SOTA CVP-MVSNet \cite{b6}~(0.292 vs 0.296). The reason for this is that the DTU dataset inherently falls under the close-range category and consists of highly heterogeneous and fine-grained object surfaces. Our ADIA module addresses this challenge by employing a Z-score mechanism, allowing it to allocate more depth hypothesis planes in the finer regions for accurate depth estimation. The qualitative results of our method compared to above methods in Fig.\ref{fig:Figure 5} also convince the above observations. Notably, our model produces more precise point cloud reconstructions on object surfaces, as indicated by the outlying rectangle in Fig.\ref{fig:Figure 5}. Moreover, our model also ranks first in Overall performance~(0.313, $ 1^{st} $) among state-of-the-art methods, e.g. UniMVSNet \cite{b42}, NP-CVP-MVSNet \cite{b52}, and Prior-Net \cite{b61}. 

\subsection{Performance on large-scale practical scenes benchmark}
Large-scale represents that the shooting distance between the camera and the scene is relatively far. Therefore, we adopted the benchmark composed of camera-captured or synthetic views in wild environments to evaluate the performance of our model on the large-scale practical scenes benchmarks, e.g. Tanks and Temples benchmark, BlendedMVS benchmark, etc.

\begin{table*}[!t]
    \centering
    \LARGE
    \resizebox{14cm}{!}{
        \begin{tabular}{c|cccccccc|ccc}
        \toprule
            Methods & Fam. $\uparrow$ & Fra. $\uparrow$ & Hor. $\uparrow$ & Lig. $\uparrow$ & M60 $\uparrow$ & Pan. $\uparrow$ & Pla. $\uparrow$ & Tra. $\uparrow$ & $F_{1}$-score $\uparrow$ & Precision $\uparrow$ & Recall $\uparrow$\\ 
            \midrule
            MVSNet \cite{b1} & 55.99 & 28.55 & 25.07 & 50.79 & 53.96 & 50.86 & 47.90 & 34.69 & 43.48 & 40.23 & 49.70\\  
            R-MVSNet \cite{b2} & 69.96 & 46.65 & 32.59 & 42.95 & 51.88 & 48.80 & 52.00 & 42.38 & 48.40 & 43.74 & 57.60\\ 
            CasMVSNet \cite{b4} & 76.37 & 58.45 & 46.26 & 55.81 & 56.11 & 54.06 & 58.18 & 49.51 & 56.84 & 47.62 & 74.01\\ 
            CVP-MVSNet \cite{b6} & 76.5 & 47.74 & 36.34 & 55.12 & 57.28 & 54.28 & 57.43 & 47.54 & 54.03 & 51.41 & 60.19\\ 
            UCS-Net \cite{b11} & 76.09 & 53.16 & 43.03 & 54.00 & 55.60 & 51.49 & 57.38 & 47.89 & 54.83 & 46.66 & 70.34\\ 
            $D^{2}$HC-RMVSNet \cite{b38} & 74.69 & 56.04 & 49.42 & 60.08 & 59.81 & 59.61 & 60.04 & 53.92 & 59.20 & 49.88 & 74.08\\ 
            AA-RMVSNet \cite{b39} & 77.77 & 59.53 & 51.53 & \textbf{64.02} & 64.05 & 59.47 & 60.85 & 55.50 & 61.51 & 52.68 & 75.69\\ 
            EPP-MVSNet \cite{b50} & 77.86 & 60.54 & 52.96 & 62.33 & 61.69 & 60.34 & \textbf{62.44} & 55.30 & 61.68 & 53.09 & 75.58\\ 
            ADR-MVSNet \cite{b62} & 72.61 & 52.81 & 37.43 & 53.46 & 51.88 & 48.71 & 59.11 & 43.76 & 52.47 & 46.62 & 63.94 \\
            IterMVS \cite{b51} & 76.12 & 55.80 & 50.53 & 56.05 & 57.68 & 52.62 & 55.70 & 50.99 & 56.94 & 47.53 & 74.69\\ 
            UniMVSNet \cite{b42} & \textbf{81.20} & 66.43 & 53.11 & 63.46 & \textbf{66.09} & \textbf{64.84} & 62.23 & \textbf{57.53} & \textbf{64.36} & \textbf{57.54} & 73.82\\ 
            Prior-Net \cite{b61} & 79.02 & 60.93 & 51.65 & 60.52 & 61.78 & 56.19 & 61.37 & 53.59 & 60.63 & 50.80 & 76.20 \\
            \midrule
            ARAI-MVSNet & 79.48	& \textbf{66.83} & \textbf{54.15} & 59.56 & 58.58 & 57.38 & 56.51 & 56.27 & 61.09 & 50.42 & \textbf{79.90} \\
        \bottomrule
        \end{tabular}
    }
    \caption{Generalization results on the TnT. The left part lists our $F_{1}$-scores on each scene. The right part lists the mean values of $F_{1}$-score, Precision and Recall.}
    \label{tab:Table.2}
\end{table*}

\begin{table}[!t]
    \centering
    \large
    \resizebox{9.5cm}{!}{
    \begin{tabular}{c|c |cccccc}
    \toprule
        Methods & Mean & Aud. & Bal. & Cou. & Mus. & Pal. & Tem. \\
        \midrule
        CasMVSNet \cite{b4} & 31.12 & 19.81 & 38.46 & 29.10 & 43.87 & 27.36 & 28.11 \\
        AA-RMVSNet \cite{b39} & 33.53 & 20.96 & 40.15 & 32.05 & 46.01 & 29.28 & 32.71 \\
        IDCF \cite{b63} & 32.28 & 23.66 & 33.01 & 36.05 & 42.10 & 23.95 & 34.92 \\
        EPP-MVSNet \cite{b50} & 35.72 & 21.28 & 39.74 & 35.34 & \textbf{49.21} & 30.00  & 38.75 \\
        ADR-MVSNet \cite{b62} & 26.75 & 20.06 & 26.60 & 27.92 & 35.28 & 26.85 & 23.77 \\
        TransMVSNet \cite{b56} & 37.00 & 24.84 & \textbf{44.59} & 34.77 & 46.49 & 34.69 & 36.62 \\
        Prior-Net \cite{b61} & 34.61 & \textbf{26.99} & 40.46 & 30.76 & 47.81 & 28.96 & 32.71 \\
        \midrule
        Our & \textbf{38.68} & 26.13 & 43.01 & \textbf{38.63} & 48.88 & \textbf{35.39} &  \textbf{40.01} \\
    \bottomrule
    \end{tabular}
    }
    \caption{The $F_{1}$ score on Tanks and Temple advanced.}
    \label{tab:Table.13}
\end{table}

\noindent \textbf{Evaluation on Tanks and Temples benchmark:} We executed a test on Tanks and Temples to evaluate the effectiveness of our model in large-scale dataset. We adopted the same dynamic geometric consistency checking strategies as $D^{2}$HC-RMVSNet \cite{b38}, and chose an appropriate mask threshold for each scene. The quantitative results in Table.\ref{tab:Table.2} indicates that our ARAI-MVSNet achieves competitive performance among all listed MVS methods and obtains SOTA results in some scenes. Our findings are as follows: our method achieves the best Recall score (79.90) and competitive results in terms of $F_{1}$-score and Precision. Notably, our ARAI-MVSNet improves the Recall score from 75.69 in AA-RMVSNet \cite{b39} to 79.90, from 63.94 in ADR-MVSNet \cite{b62} to 79.90, from 76.20 in Prior-Net \cite{b61} to 79.90, and from 73.82 in UniMVSNet \cite{b42} to 79.90 ($1^{st}$), while maintaining comparable $F_{1}$-score and precision with all other MVS methods listed. Moreover, our method also obtain the highest $F_{1}$-score in Family~(66.83) and Horse~(54.15) compared with all listed methods. 
The Tanks and Temples benchmark dataset comprises large-scale scenes, wherein the objects of interest occupy only a fraction of the entire scene. Our proposed Adaptive Depth Range Prediction (ADRP) module, equipped with an adaptive boundaries adjustment mechanism, enables accurate prediction of the all-pixel depth range by zooming in on the scene, resulting in more comprehensive reconstruction results. We conclude that the improvement in Recall is due to the mechanism of the ADRP.

We also executed the evaluation on TnT advanced dataset~(Table \ref{tab:Table.13}). Our findings are as follows: our model achieves the best Mean $F_{1}$-score than all listed methods. In particular, our model improves the $F_{1}$-score from 33.53 in AA-RMVSNet \cite{b39} to 38.68, from 32.28 in IDCF \cite{b63} to 38.68, from 26.75 in ADR-MVSNet \cite{b62} to 38.68, from 34.61 in Prior-Net \cite{b61} to 38.68, and from 37.00 in TransMVSNet \cite{b56} to 38.68 ($1^{st}$). The $F_{1}$-score is calculated from both Precision and Recall. Therefore, achieving SOTA $F_{1}$-score means that our model performs exceptionally well in both Precision and Recall. The outstanding results are mainly attributed to the superior performance of our proposed ADIA module in Precision and the excellent performance of our ADRP module in Recall. Besides, our method can obtain the highest results on several scenes, i.e. Courtroom, Palace and Temple. 

\noindent \textbf{Evaluation on BlendedMVS benchmark:} As shown in Table.\ref{tab:Table.10_11} left, we tested our model on BlendedMVS \cite{b33} validation set to further demonstrate the generalizability and flexibility of ARAI-MVSNet. We can observe that our method achieves the highest scores in $e_{1}$ and $e_{3}$. Specially, we improve the $e_{1}$ from 12.66 in EPP-MSVNet \cite{b50} to 7.91 and from 9.35 in UniMVSNet \cite{b42} to 7.91. Moreover, we also improve the $e_{3}$ from 6.20 in EPP-MSVNet \cite{b50} to 2.95 and from 3.25 in UniMVSNet \cite{b42} to 2.95. $e_{1}$ and $e_{3}$ represent the proportion in $\%$ of pixels with depth error larger than 1 and larger than 3, the SOTA performances on $e_{1}$ and $e_{3}$ confirm the advantage of our ADIA module as ADIA can estimate more accurate depth values by reallocating the depth planes. Our ADRP module pays more attention to the depth range of the foreground but may ignore the depth range of the background, thus, the calculation of EPE may be affected by outliers. Therefore, our EPE is slightly lower than UniMVSNet \cite{b42}~(0.67 vs 0.62), but higher than all other models.

\begin{table}[!t]
    \centering
    \resizebox{11.5cm}{!}{
    \begin{tabular}{c|ccc||c|ccccc}
    \toprule
        Methods & $e_{1}$ $\downarrow$ & $e_{3}$ $\downarrow$ & EPE $\downarrow$ & Methods & Acc $\uparrow$ & Comp $\uparrow$ & $F_{1}$-score $\uparrow$ \\
        \midrule
        MVSNet \cite{b1}  & 21.98 & 8.32 & 1.49 & Gipuma \cite{b21} & 86.47 & 24.91 & 45.18 \\
        UCS-MVSNet \cite{b11} & 14.12 & 7.33 & 1.32 & PatchmatchNet \cite{b49} & 69.71 & 77.46 & 73.12 \\
        EPP-MVSNet \cite{b50} & 12.66 & 6.20 & 1.17 & Iter-MVS \cite{b51} & 76.91 & 72.65 & 74.29 \\
        TransMVSNet \cite{b56} & 8.32 & 3.62 & 0.73 & PatchMatch-RL \cite{b49} & 74.48 & 72.06 & 72.38 \\
        UniMVSNet \cite{b42} & 9.35 & 3.25 & \textbf{0.62} & Iter-MVS-RL \cite{b51} & 84.73 & 76.49 & 80.06 \\
        \midrule
        Our & \textbf{7.91} & \textbf{2.95} & 0.67 & Our & \textbf{92.12} & 78.73 & \textbf{84.90} \\
    \bottomrule
    \end{tabular}
    }
    \caption{\textbf{Left part}: The results on BlendedMVS. EPE stands for the endpoint error, which is the average $\zeta $-1 distance between the prediction and the ground truth depth; $e_{1}$ and $e_{3}$ represent the proportion in $\%$ of pixels withdepth error larger than 1 and larger than 3. \textbf{Right part}: The evaluation results on ETH 3D are obtained from Iter-MVS.}
    \label{tab:Table.10_11}
\end{table}

\noindent \textbf{Evaluation on ETH 3D benchmark:} We conducted further experiments on the large-scale scene dataset (ETH 3D \cite{b34}) to validate the performance of ARAI-MVSNet in the realistic wild environment. We can observe that our method obtain the highest Acc and F1-score on ETH 3D (as shown in Table.\ref{tab:Table.10_11}, right)). In particular, we have the following observations: our ARAI-RMVSNet improves the Acc from 74.48 of PatchMatch-RL \cite{b49} to 92.12 and 84.73 of Iter-MVS-RL \cite{b42} to 92.12. And our ARAI-RMVSNet also improves the $F_{1}$-score from 72.38 of PatchMatch-RL \cite{b49} to 84.90 and 80.06 of Iter-MVS-RL \cite{b42} to 84.90. Our model achieves SOTA performance on Acc when benefitting from the accurate depth planes reallocated by ADIA. However, we have noticed that different lighting conditions in the ETH 3D dataset, unlike DTU and BlendedMVS datasets used for training, may cause our ADRP module to estimate an incorrect all-pixel depth range due to dark lighting. This incorrect estimation can adversely impact the reconstruction results on Comp. Moreover, the SOTA Acc and competitive Comp of our model enable it to achieve SOTA $F_{1}$-score. These indicate that our method significantly improves overall reconstruction quality and demonstrates competitive generalization ability compared to most deep learning-based MVS methods.

\begin{table*}[!t]
    \centering
    \huge
    \resizebox{14cm}{!}{
        \begin{tabular}{c|cccccc|cc}
        \toprule
            Methods & Input Size & View & Num Depth & Acc(mm) $\downarrow$ & Comp(mm) $\downarrow$ & Overall(mm) $\downarrow$ & GPU Mem(MB) & Time(s)\\ 
            \midrule
            CasMVSNet \cite{b4} & 1280 $\times$ 960   & 7 & [64,16,8] & 0.3508 & 0.4039 & 0.37735 & 4939MB & 0.64s\\ 
            CVP-MVSNet \cite{b6} & 1280 $\times$ 960   & 7 & - & 0.3023 & 0.4256 & 0.36395 & 6327MB  & 1.51s\\ 
            UCS-Net \cite{b11} & 1280 $\times$ 960   & 7 & [64,16,8] & 0.3609 & 0.3764 & 0.36865 & \textbf{4889MB} & 0.77s\\ 
            UniMVSNet \cite{b42} & 1280 $\times$ 960   & 7 & [64,16,8] & 0.3772 & \textbf{0.2912} & 0.3342 & 6441MB & 0.88s\\ 
            \midrule
            ARAI-MVSNet & 1280 $\times$ 960   & 7 & [16,64,16,8] & \textbf{0.2924} & 0.3342 & \textbf{0.3133} & 5386MB & \textbf{0.61s}\\
        \bottomrule
        \end{tabular}
    }
    \caption{Performance comparisons on the DTU val set. "Time" is the inference time.}
    \label{tab:Table.2.1}
\end{table*}

\subsection{Comparison of the Models Efficiency}
In this section, we firstly compared the inference time with several multi-stage MVS methods on the DTU dataset. The results in Table.\ref{tab:Table.2.1} indicate that our method produces accurate point clouds with lower run-time than previous methods, with reductions of 0.9s, 0.16s, and 0.27s compared to CVP-MVSNet \cite{b6}, UCS-Net \cite{b11}, and UniMVSNet \cite{b42}, respectively. Our model also has competitive memory usage~(5386MB) compared to mainstream methods \cite{b4,b6,b11,b42}. Although our four-stage framework has slightly higher memory usage than UCS-Net, it performs better in other efficiency metrics. Compared to UniMVSNet, which has the best Comp score of 0.2912, our model achieves better scores in terms of Acc and Overall performance and has a faster inference time. The non-parametric mechanism in our ADIA module significantly improves the inference time. \revise{Furthermore, we additionally compared our method with some superior Transformer-based methods~(e.g. MVSFormer \cite{b70}, WT-MVSNet \cite{b71}). The quantitative results in Table \ref{tab:Table.R1.5} reveal that our method achieves better or comparable results~(highest Recall: 79.90) with lower memory and time consumption than Transformer-based methods, with reductions of 61\% and 18\% in memory and 8\% and 44\% in time compared to MVSFormer \cite{b70} and WT-MVSNet \cite{b71}. We attribute our model's efficiency superiority to two main reasons. Firstly, methods like MVSFormer, WT-MVSNet, etc., being Transformer-based, inherently demand more computation and memory resources compared to ARAI-MVSNet, which adopts a simpler and more lightweight CNN-based framework. Secondly, the ADIA module proposed in our work utilizes a simple and non-parametric mechanism~(Z-score), introducing minimal additional memory and computational overhead while still delivering performance improvements.}

\begin{table*}[!t]
    \centering
    \footnotesize
    \resizebox{13cm}{!}{
    \begin{tabular}{c|ccc||cccc}
    \toprule
        Methods & $F_{1}$-score $\uparrow$ & Precision $\uparrow$ & Recall $\uparrow$ & Input Size & View & Memory (MB) $\downarrow$ & Time (s) $\downarrow$\\ 
        \midrule
        MVSFormer & \textbf{66.41} & \textbf{60.77} & 74.33 & 864 × 1152 & 7 & 11662MB & 0.614s \\
        WT-MVSNet & 65.34 & 56.09 & 79.04 & 864 × 1152 & 7 & 5549MB & 0.978s \\
        ARAI-MVSNet~(Ours) & 61.09 & 50.42 & \textbf{79.90} & 864 × 1152 & 7 & \textbf{4562MB} & \textbf{0.543s} \\
    \bottomrule
    \end{tabular}
    }
    \caption{Quantitative results on TnT (Left) and efficiency results on DTU (Right). The results of WT-MVSNet are from the official paper since the code is not open source.}
    \label{tab:Table.R1.5}
\end{table*}

\subsection{Quantitative analysis of depth estimation results}
The accuracy of the depth map directly affects the final reconstructed point cloud quality. Our model has two novel modules, ADRP and ADIA. Specifically, ADRP predicts an adaptive depth range to zoom in a scene, while ADIA achieves adaptive depth interval partition by reallocating the depth hypothesis planes through offset learning, resulting in more accurate depth values. The effectiveness of these modules were also evaluated based on the depth estimate results, as follows: 



\begin{table}[!t]
    \centering
    \footnotesize
    \resizebox{8cm}{!}{
    \begin{tabular}{c|ccc}
    \toprule
         Methods & Mean AOG $\uparrow$ & Mean AOS $\uparrow$ & F-score $\uparrow$ \\
        \midrule
        $\mathbf{D}_{128}$ & 55.56\% & 57.46\% & 56.49\% \\
        $\mathbf{D}_{256}$ & \textbf{98.58}\% & 56.33\% & 71.69\% \\
        $\mathbf{D}_{adrp}$~(Ours) & 90.88\% & \textbf{79.27}\% & \textbf{84.68}\% \\

    \bottomrule
    \end{tabular}
    }
    \caption{The overlap rate of predicted and fixed all-pixel depth range. F-score is the harmonic mean of AOG and AOS.} 
    \label{tab:Table.4.2}
\end{table}

\noindent \textbf{Comparison with depth range overlap rate:} In this section, we compared the overlap rate of different all-pixel depth ranges predicted by ADRP, with depth hypothesis planes set to 128 and 256, with the ground truth all-pixel depth range. The results are shown in Table. \ref{tab:Table.4.2}. From the table, we can observe that our predicted adaptive all-pixel depth range can cover objects more comprehensively than the fixed all-pixel depth range while using fewer depth hypothesis planes. Suppose $\mathbf{D}_{adrp}$ is the predicted all-pixel depth range $\mathbf{D}_{adrp}$ which is obtained by ADRP, $\mathbf{D}_{128}(\mathbf{D}_{num}=128)$ and $\mathbf{D}_{256}(\mathbf{D}_{num}=256)$ are the two default all-pixel depth range in most models\cite{b1,b2}. To evaluate whether our ADRP module can obtain a more accurate all-pixel depth range for the scene reconstruction, two metrics to measure the overlap between the adopted all-pixel depth range and the GT all-pixel depth range $\mathbf{D}_{gt}$ obtained by the GT depth map are defined:

\begin{equation}
    \label{eq:equation_14}
    \text{AOG} = \frac{|\mathbf{D}_{gt} \cap  \mathbf{D}_{i}|}{|\mathbf{D}_{gt}|}, \quad
    \text{AOS} = \frac{| \mathbf{D}_{gt} \cap \mathbf{D}_{i} |}{|\mathbf{D}_{i}|} \\
\end{equation}

\noindent where $ \mathbf{D}_{i} \in [\mathbf{D}_{adrp}, \mathbf{D}_{128}, \mathbf{D}_{256}] $, $\cap$ represents the intersection operation and $ | \cdot | $ is the length of the depth range. A higher AOG represents a higher overlap rate with the GT all-pixel depth range, while a higher AOS denotes less waste in the all-pixel depth range.

The mean results on the DTU evaluation set are shown in Tab. \ref{tab:Table.4.2}. The results indicate that by setting a smaller number of depth hypothesis planes $\mathbf{D}_{128}$, the fixed all-pixel depth range ~(e.g. $[425, 425 + 2.5 \times 128]$) cannot cover enough effective space for the scene~(reflected on MAOG=55.56\% and MAOG=57.46\%), resulting in low performance. Meanwhile, even if the number of depth hypothesis planes is set to be a larger value~(e.g. $\mathbf{D}_{num}=256$) to enlarge the all-pixel depth range to cover the GT all-pixel depth range~(MAOG=98.58\%), the overhead of the model will increase and lead to the wastage of the depth planes, i.e., only 56.33\% of the depth planes are actually useful. On contrary, our model covers 90.88\% of the GT scene and with more percentage of depth hypothesis planes being utilized, i.e., MAOS=79.27\%, which proves that our ADRP module can predict an adaptive and accurate all-pixel depth range close to the GT all-pixel depth range.

\begin{table}[!t]
    \centering
    \footnotesize
    \resizebox{7.5cm}{!}{
    \begin{tabular}{c|ccc|c}
    \toprule
        \multirow{2}*{Methods} & \multicolumn{3}{c|}{Modules} & \multirow{2}*{$P_{numD}$ $\uparrow$} \\ \cline{2-4} & ASPFNet & ADRP  & ADIA \\
        \midrule
        - ADIA & \checkmark & \checkmark & & 32.97\% \\
        + ADIA & \checkmark & \checkmark & \checkmark  & \textbf{61.76}\% \\
    \bottomrule
    \end{tabular}
    }
    \caption{The total number of pixels with predicted values close to the GT depth value.} 
    \label{tab:Table.4.1}
\end{table}

\noindent \textbf{Comparison with total accurate number of pixels:} As we mentioned in aforementioned section, the calculation of pixel-wise depth range in the current phase involves using the variance of the previous phase's depth values, as defined in Eq. \eqref{eq:equation_5}, after achieving the equal interval partition. However, employing the equal depth interval partition strategy may lead to inadequate depth hypothesis planes due to the heterogeneous nature of object surfaces, resulting in imprecise depth estimation outcomes. Therefore, in this section, we conducted a comparative experiment to verify the effectiveness of our ADIA. Based on the pixel-wise depth range, we have examined two scenarios: one where we employed the adaptive variable interval partition and another employed the equal interval partition. Then we calculated the total number of pixels where the predicted depth value was close to the GT depth value to validate whether our ADIA can achieve more accurate depth estimation for each pixel. Specifically, we calculated the distance~($D_{e}$) between the predicted depth value and the GT depth value for all pixels in DTU evaluation set, and displayed the percentage of the number of pixels~($P_{numD}$) which were close to the ground truth~(i.e., $D_{e} \le 1e-1$) in Tab.\ref{tab:Table.4.1}. The results demonstrate that applying the adaptive variable interval partition improves the $P_{numD}$ by a large margin, which indicates that our ADIA module has estimated more accurate depth values for each pixel by setting adaptive depth hypothesis planes. In conclusion, achieving adaptive variable interval partition based on the pixel-wise depth range is imperative.

\subsection{Ablation Experiments}
In this section, we set our baseline using a general feature pyramid network and 3D-UNet for cost volume regularization. We then combined our proposed modules and demonstrated their effectiveness with different combinations of modules in Table \ref{tab:Table.4}, left. We also conducted ablation experiments for ADRP and ADIA in real scenes (TnT) in Table \ref{tab:Table.4}, right.

\noindent \textbf{Effectiveness of ADRP:} As shown in Table.\ref{tab:Table.4}(left), we can observe that the adaptive all-pixel depth range obtained by our ADRP can effectively improve the performance on DTU dataset~(Comp: from 0.3982 to 0.3342). As shown in Table.\ref{tab:Table.4}(right), we can observe that the Recall is improved from 72.63 to 79.90 with ADPR on the Tanks and Temples intermediate benchmark. These results validate the effectiveness of our proposed ADRP.

\begin{table*}[!t]
    \centering
    \LARGE
    \resizebox{12cm}{!}{
    \begin{tabular}{c|ccc|ccc|ccc}
    \toprule
        \multirow{2}*{Methods} & \multicolumn{3}{c|}{Modules} & \multirow{2}*{Acc$\downarrow$} & \multirow{2}*{Comp$\downarrow$} & \multirow{2}*{Overall$\downarrow$} & \multirow{2}*{$F_{1}$-score $\uparrow$} & \multirow{2}*{Precision $\uparrow$} & \multirow{2}*{Recall $\uparrow$} \\ \cline{2-4} & ASPFNet & ADRP & ADIA \\
        \midrule
        - ADIA & \checkmark & \checkmark &  & 0.3667 & 0.3693 & 0.3680 & 57.33 & 45.48 & 77.53 \\
        - ADRP & \checkmark &  & \checkmark  & 0.3021 & 0.3982 & 0.3502 & 58.16 & 48.50 & 72.63\\ 
        - ASPFNet &  & \checkmark & \checkmark & 0.3143 & 0.3532 & 0.3338 & - & - & - \\
        \midrule
        ARAI-MVSNet & \checkmark & \checkmark & \checkmark & \textbf{0.2924} & \textbf{0.3342} & \textbf{0.3133} & \textbf{61.09} & \textbf{50.42} & \textbf{79.90} \\
    \bottomrule
    \end{tabular}
    }
    \caption{\textbf{Left part}: Ablation results on DTU evaluation set. The left part of table lists the mean values of Acc, Comp and Overall. \textbf{Right part}: Ablation results on TnT dataset. The right part of table lists the mean values of $F_{1}$-score, Precision and Recall.}
    \label{tab:Table.4}
\end{table*}

\noindent \textbf{Effectiveness of ADIA:} As shown in Table.\ref{tab:Table.4}(left), we can observe that the adaptive pixel-wise depth interval reallocated by our ADIA module can significantly improves the Acc from 0.3667 to 0.2924 on DTU dataset. The significant enhancement in performance can be attributed to the utilization of Z-score distribution-based offset calculation. In the case of small-scale scenes, the two-stage ADIAs effectively explores precise depth values within a reduced full-pixel depth range, leading to a higher level of accuracy in the depth map. Furthermore, the results exhibited in Table \ref{tab:Table.4}(right), demonstrate an improvement in Precision from 45.48 to 50.42 on on the Tanks and Temples intermediate benchmark. By leveraging the accurate all-pixel depth range predicted by ADPR for large-scale scenes, our model demonstrates an improvement in Precision, even with the utilization of two-stage ADIAs. These results corroborate the effectiveness of our proposed ADIA.

\noindent \textbf{Effectiveness of ASPFNet:} Tab.\ref{tab:Table.4}(left) shows that ASPFNet uses multi-path dilated convolutions to extract multi-scale contextual features, resulting in a larger receptive field and improved reconstruction quality (Acc: 0.3143 to 0.2924 and Comp: 0.3532 to 0.3342) on DTU dataset.

\begin{table}[!t]
    \centering
    \footnotesize
    \resizebox{12cm}{!}{
    \begin{tabular}{c|ccc|ccc}
    \toprule
        Methods & Acc. (mm) $\downarrow$ & Comp. (mm) $\downarrow$ & Overall (mm) $\downarrow$ & $F_{1}$-score $\uparrow$ & Precision $\uparrow$ & Recall $\uparrow$ \\ \hline
        ARAI-MVSNet &  0.2924 & \textbf{0.3342} & \textbf{0.3133} & \textbf{61.09} & \textbf{50.42} & \textbf{79.90} \\
        replace ADRP & \textbf{0.2901} & 0.3402 & 0.3152 & 58.46 & 48.11 & 74.48 \\
    \bottomrule
    \end{tabular}
    }
    \caption{The ablation study to verify the effectiveness and superiority of our ADRP.}
    \label{tab:Table.R1.3}
\end{table}

\subsection{High-Level Analysis}
In this section we have provided some ablation experiments for high-level analysis of the proposed ADRP module and ADIA module.

\noindent \textbf{High-Level Analysis of ADRP:} To verify the superiority of the image-level uniform adjustment of our proposed ADRP, we have replaced the ADRP with ADIA in stage 2 to execute an ablation study. From Table \ref{tab:Table.R1.3}, we can observe that replacing ADRP with ADIA in stage 2 achieves similar Acc results (0.2901 vs 0.2924) on the DTU benchmark, but it performs slightly worse in terms of Comp and Overall indicators compared to ARAI-MVSNet~(Comp: 0.3402 vs 0.3342, Overall:0.3152 vs 0.3133). While for the TnT metric, ARAI-MVSNet achieves better performance in all metrics ($F_{1}$-score: 61.09 vs 58.46, Precision: 50.42 vs 48.11, Recall: 79.90 vs 74.48). These indicate the superiority of our ADRP, namely, predicting a reliable and accurate all-pixel depth range of the scene as a base for subsequent ADIA is effective, especially in large-scale scenes. As for the similar Acc on DTU benchmark, the reason is that targets in large scale scales have varying sizes which can't make full use of an initial setting all-pixel depth range. While in close-range scale, the targets tend to have similar sizes, which are more likely to make full use of an initial setting all-pixel depth range.

\begin{table}[!t]
    \centering
    \footnotesize
    \resizebox{9.5cm}{!}{
    \begin{tabular}{c|ccc}
    \toprule
        Methods & Acc. (mm) $\downarrow$ & Comp. (mm) $\downarrow$ & Overall (mm) $\downarrow$ \\ \hline
        ADIA + Z-score distribution & \textbf{0.2924} & \textbf{0.3342} & \textbf{0.3133} \\
        ADIA + Linear distribution & 0.3345 & 0.3481 & 0.3413 \\
    \bottomrule
    \end{tabular}
    }
    \caption{The quantitative experiment of two different distributions in ADIA.}
    \label{tab:Table.R1.4}
\end{table}

\begin{figure}[!t]
\centering
\includegraphics[width=1\textwidth]{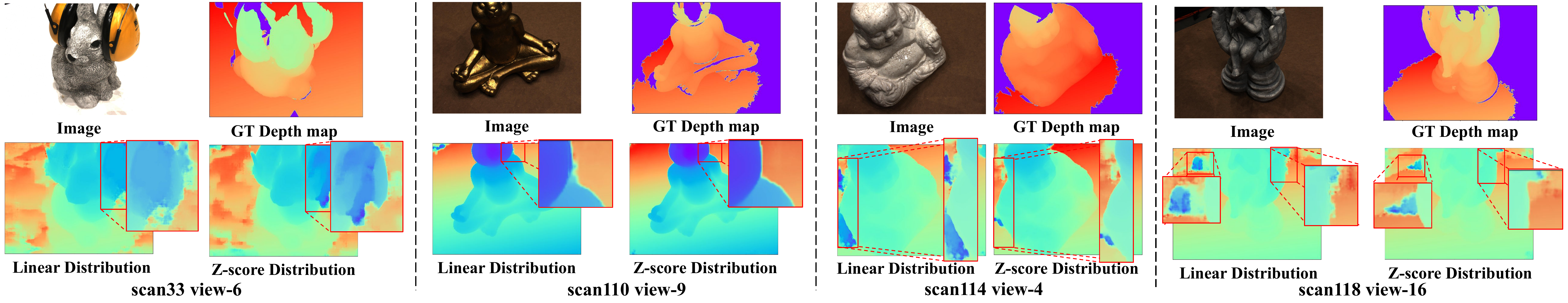}
\caption{Quantitative results. We compared the details of depth maps from four scans obtained by Linear distribution and Z-score distribution separately.}
\label{fig:z-score_vs_linear}
\end{figure}

\noindent \textbf{High-Level Analysis of ADIA:} To verify the superiority of the Z-score distribution-based offset calculation of our ADIA, we have conducted an ablation study by replacing the Z-score distribution in ADIA with a simpler linear distribution~(i.e. removing $\sigma$ from Eq. \eqref{eq:equation_7}). As shown in Table \ref{tab:Table.R1.4}, we can observe that ADIA performs better in terms of Acc and Comp when using the Z-score distribution compared to the Linear distribution, especially in Acc~(0.2924 vs 0.3345). We attribute this improvement to the Z-score, which allows for the allocation of more depth hypothesis planes around potential ground truth depth values, which enables the estimation of more accurate depth values. As shown in Eq. \eqref{eq:equation_7}, the value of $\sigma$ is dynamically adjusted as the network continues to optimize, facilitating adaptive and high-precision adjustment of the depth plane positions for accurate depth estimation. \revise{Figure \ref{fig:z-score_vs_linear} presents a qualitative comparison between ADIA+Linear distribution and ADIA+Z-score distribution for the depth map of a view from four scans on the DTU benchmark. The depth maps estimated using Z-score distribution exhibit superior overall quality compared to those obtained with Linear distribution. Particularly, examining the red boxes in the four scenes reveals that our method not only achieves better depth maps in terms of overall image quality but also provides more accurate depth estimation in detail. This improvement is attributed to the Z-score distribution, which assigns more depth hypothesis planes around potential depth values, thanks to the dynamic $\sigma$, leading to more precise depth estimation.}

\section{Limitations and discussion}
Although our model achieves better or comparable performance than most of the state-of-the-art methods on the four benchmarks \cite{b28, b32, b33, b34}, it has several limitations: (1)~Most extensive multi-view stereo (MVS) datasets, such as DTU and TnT, rely on numerous overlapped views for high-quality point cloud reconstruction. However, the performance of our model may decrease with fewer views, leading to poorer reconstruction results.  This problem can be alleviated by integrating a module into MVSNet that can generate synthetic views based on limited real views, such as the neural radiance fields. (2)~In current large-scale practical scene datasets, such as TnT, camera parameters are often derived from real outdoor scenes, and may not be precisely accurate. While MVSNet uses differentiable homography warping based on the camera parameters to construct the cost volume, and inaccurate camera parameters may lead to lower depth map estimation quality. To address this issue, the incorporation of a module capable of optimizing the camera parameters of inaccurate views may provide accurate camera parameters, which in turn can enable us to solve the dense structure from motion (SfM) problem. (3)~With the emergence of advanced modules like Transformers, certain Transformer-based methods have gained prominence, such as MVSFormer \cite{b70}. While they exhibit superior performance in specific indicators, such as $F_{1}$-score on the TnT, this advantage comes at the cost of increased computational and memory requirements. Despite this, our method outperforms others in terms of Recall. In the future, we consider integrating these modules into more sophisticated multi-stage baselines but with efficiency into consideration. This integration is expected to yield improved or competitive performance across various metrics efficiently.

\section{Conclusion}
In this paper, we propose a novel multi-stage coarse-to-fine framework ARAI-MVSNet for high-quality reconstruction. Our ADRP module leverages the reference image and depth map of the previous stage to compute range adjustment parameters to achieve adaptively depth range adjustment, which can effectively zoom in an accurate all-pixel depth range of the scene. Further, the ADIA module can reallocate the pixel-wise depth interval distribution to obtain more accurate depth values by utilizing the Z-score distribution for the adaptive variable interval partition. The experimental results demonstrate that our proposed method achieves state-of-the-art performance and exhibits competitive generalization ability compared to all listed methods. Specifically, our approach outperforms other pioneer works on the DTU dataset and obtains the highest recall score and $F_{1}$-score on the Tanks and Temples intermediate and advanced datasets. Additionally, our method achieves the lowest $e_{1}$ and $e_{3}$ scores on the BlendedMVS dataset and the highest accuracy and $F_{1}$-score on the ETH 3D dataset. These significant improvements confirm the advantages of our proposed modules. In the future, we plan to explore (1) combining ARAI-MVSNet with neural radiance fields to enable high-precision reconstruction with fewer views in MVS datasets, (2) incorporating dense bundle adjustment to jointly optimize camera poses and depth maps to enhance the robustness and accuracy of ARAI-MVSNet.









\bibliographystyle{elsarticle-num}
\bibliography{mybibfile}

\end{document}


\maketitle

In this supplementary material, we introduce more details about (1) the network architecture~(Sec. \ref{sec:A}), (2) the quantitative results~(Sec. \ref{sec:B}), (3) the qualitative results~(Sec. \ref{sec:C}), and (4) the reconstruction results~(Sec. \ref{sec:D}).

\section{More Architecture Details}
\label{sec:A}
In this section, we introduce the details about the Atrous Spatial Pyramid Feature Extraction Network (ASPFNet), cost volume regularization and differentiable homography warping.
\subsection{Network Architecture of ASPFNet}
\begin{figure}[!ht]
\centering
\includegraphics[width=0.65\textwidth]{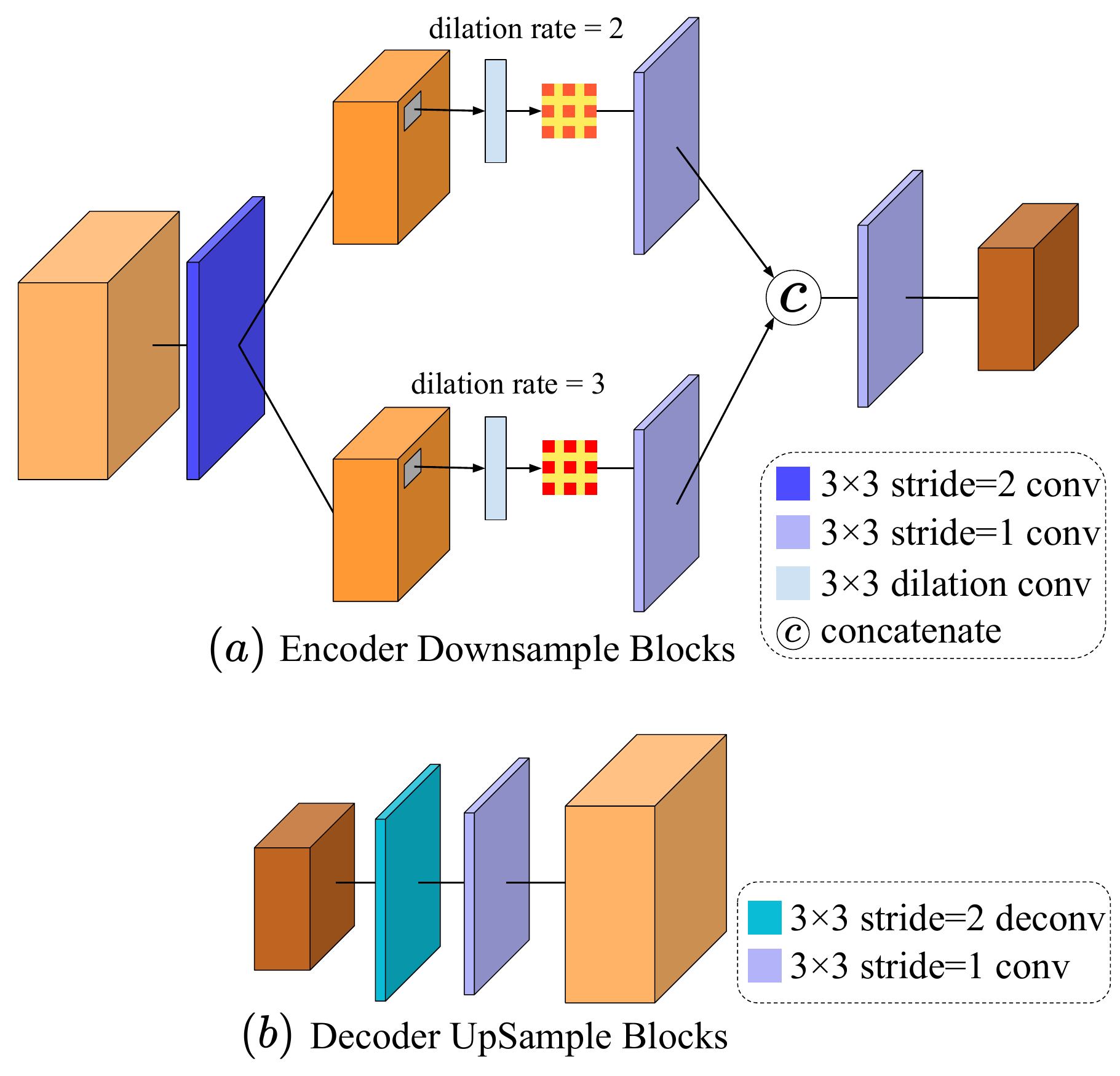}
\caption{Illustration of Encoder Downsample Block~(EDB) and Decoder UpSample Block~(DUB).}
\label{fig:Figure 1}
\end{figure}

The ASPFNet is an encoder-decoder architecture that is used for feature extraction. It is composed of four Encoder Downsample Blocks~(EDB), four Decoder UpSample Blocks~(DUB), and skip connections. For each EDB, we construct two-way feature fusion branches by using two dilated convolutions with different dilation rates~(2,3) to achieve high-level information extraction and contextual information fusion. Then the outputs of two-way feature fusion branches will be concatenated. And we use a $1\times 1$ convolution to adjust the dimension of channels after the concatenation. The schematic of EDB is shown in Fig.\ref{fig:Figure 1}.(a). For each DUB, we use a deconvolution and an ordinary convolution cascade to achieve feature map upsampling, and the details of DUB are shown in Fig.\ref{fig:Figure 1}.(b). The input size and output size of network are listed in Table.\ref{tab:Table.1}

\begin{table}[!ht]
    \centering
    \footnotesize
    \resizebox{7.5cm}{!}{
    \begin{tabular}{c|c|c}
    \toprule
        \midrule
        \multicolumn{3}{c}{\textbf{Input Images:} $3 \times H \times W $} \\ \midrule
        Layer Description & Input Size & Output Size \\ \midrule
        \multicolumn{3}{c}{\textbf{Downsample Layers}} \\ \midrule
        EDB 1 & $ 3 \times H \times W $ & $ 8 \times H \times W $ \\
        EDB 2 & $ 8 \times H \times W $ & $ 16 \times \frac{H}{2} \times \frac{W}{2} $ \\
        EDB 3 & $ 16 \times \frac{H}{2} \times \frac{W}{2} $ & $ 32 \times \frac{H}{4} \times \frac{W}{4} $ \\
        EDB 4 & $ 32 \times \frac{H}{4} \times \frac{W}{4} $ & $ 64 \times \frac{H}{8} \times \frac{W}{8} $ \\
        \midrule
        \multicolumn{3}{c}{\textbf{Upsample Layers}} \\ \midrule
        DUB 1 & $ 64 \times \frac{H}{8} \times \frac{W}{8} $ & $ 64 \times \frac{H}{8} \times \frac{W}{8} $ \\
        DUB 2 & $ 32 \times \frac{H}{4} \times \frac{W}{4} $ & $ 32 \times \frac{H}{4} \times \frac{W}{4} $ \\
        DUB 3 & $ 32 \times \frac{H}{4} \times \frac{W}{4} $ & $ 16 \times \frac{H}{2} \times \frac{W}{2} $ \\
        DUB 4 & $ 16 \times \frac{H}{2} \times \frac{W}{2} $ & $ 8 \times H \times W $ \\
        \midrule
    \bottomrule
    \end{tabular}
    }
    \caption{The detailed architecture of the ASPFNet.}
    \label{tab:Table.1}
\end{table}

\begin{table}[!ht]
    \centering
    \footnotesize
    \resizebox{10.5cm}{!}{
    \begin{tabular}{c|c|c}
    \toprule
        \midrule
        Stage Description & Layer Description & Output Size \\
        \midrule
        \multicolumn{3}{c}{\textbf{Input Cost Volume} $\mathbf{V}_{s_{1}}$ \text{:} $64 \times 16 \times \frac{H}{8} \times \frac{W}{8} $} \\
        \midrule
        UNet Stage 1 conv\_0 & Conv3D, 3 × 3 × 3, S1, P1 & $ 8 \times 16 \times \frac{H}{8} \times \frac{W}{8} $ \\
        UNet Stage 1 conv\_1 & Conv3D, 3 × 3 × 3, S2, P1 & $ 16 \times 8 \times \frac{H}{16} \times \frac{W}{16} $ \\
        UNet Stage 1 conv\_2 & Conv3D, 3 × 3 × 3, S2, P1 & $ 32 \times 4 \times \frac{H}{32} \times \frac{W}{32} $ \\
        UNet Stage 1 conv\_3 & Conv3D, 3 × 3 × 3, S2, P1 & $ 64 \times 2 \times \frac{H}{64} \times \frac{W}{64} $ \\
        
        UNet Stage 1 deconv\_4 & TransposeConv3D, 3 × 3 × 3, S2, P1, OP1 & $ 32 \times 4 \times \frac{H}{32} \times \frac{W}{32} $ \\
        UNet Stage 1 deconv\_5 & TransposeConv3D, 3 × 3 × 3, S2, P1, OP1 & $ 16 \times 8 \times \frac{H}{16} \times \frac{W}{16} $ \\
        UNet Stage 1 deconv\_6 & TransposeConv3D, 3 × 3 × 3, S2, P1, OP1 & $ 8 \times 16 \times \frac{H}{8} \times \frac{W}{8} $ \\
        \midrule

        \multicolumn{3}{c}{\textbf{Input Cost Volume} $\mathbf{V}_{s_{2}}$ \text{:} $32 \times 48 \times \frac{H}{4} \times \frac{W}{4} $} \\
        \midrule
        UNet Stage 2 conv\_0 & Conv3D, 3 × 3 × 3, S1, P1 & $ 8 \times 48 \times \frac{H}{4} \times \frac{W}{4} $ \\
        UNet Stage 2 conv\_1 & Conv3D, 3 × 3 × 3, S2, P1 & $ 16 \times 24 \times \frac{H}{8} \times \frac{W}{8} $ \\
        UNet Stage 2 conv\_2 & Conv3D, 3 × 3 × 3, S2, P1 & $ 32 \times 12 \times \frac{H}{16} \times \frac{W}{16} $ \\
        UNet Stage 2 conv\_3 & Conv3D, 3 × 3 × 3, S2, P1 & $ 64 \times 6 \times \frac{H}{32} \times \frac{W}{32} $ \\
        
        UNet Stage 2 deconv\_4 & TransposeConv3D, 3 × 3 × 3, S2, P1, OP1 & $ 32 \times 12 \times \frac{H}{16} \times \frac{W}{16} $ \\
        UNet Stage 2 deconv\_5 & TransposeConv3D, 3 × 3 × 3, S2, P1, OP1 & $ 16 \times 24 \times \frac{H}{8} \times \frac{W}{8} $ \\
        UNet Stage 2 deconv\_6 & TransposeConv3D, 3 × 3 × 3, S2, P1, OP1 & $ 8 \times 48 \times \frac{H}{4} \times \frac{W}{4} $ \\
        \midrule

        \multicolumn{3}{c}{\textbf{Input Cost Volume} $\mathbf{V}_{s_{3}}$ \text{:} $16 \times 16 \times \frac{H}{2} \times \frac{W}{2} $} \\
        \midrule
        UNet Stage 3 conv\_0 & Conv3D, 3 × 3 × 3, S1, P1 & $ 8 \times 16 \times \frac{H}{2} \times \frac{W}{2} $ \\
        UNet Stage 3 conv\_1 & Conv3D, 3 × 3 × 3, S2, P1 & $ 16 \times 8 \times \frac{H}{4} \times \frac{W}{4} $ \\
        UNet Stage 3 conv\_2 & Conv3D, 3 × 3 × 3, S2, P1 & $ 32 \times 4 \times \frac{H}{8} \times \frac{W}{8} $ \\
        UNet Stage 3 conv\_3 & Conv3D, 3 × 3 × 3, S2, P1 & $ 64 \times 2 \times \frac{H}{16} \times \frac{W}{16} $ \\
        
        UNet Stage 3 deconv\_4 & TransposeConv3D, 3 × 3 × 3, S2, P1, OP1 & $ 32 \times 4 \times \frac{H}{8} \times \frac{W}{8} $ \\
        UNet Stage 3 deconv\_5 & TransposeConv3D, 3 × 3 × 3, S2, P1, OP1 & $ 16 \times 8 \times \frac{H}{4} \times \frac{W}{4} $ \\
        UNet Stage 3 deconv\_6 & TransposeConv3D, 3 × 3 × 3, S2, P1, OP1 & $ 8 \times 16 \times \frac{H}{2} \times \frac{W}{2} $ \\
        \midrule

        \multicolumn{3}{c}{\textbf{Input Cost Volume} $\mathbf{V}_{s_{4}}$ \text{:} $8 \times 8 \times H \times W $} \\
        \midrule
        UNet Stage 4 conv\_0 & Conv3D, 3 × 3 × 3, S1, P1 & $ 8 \times 8 \times H \times W $ \\
        UNet Stage 4 conv\_1 & Conv3D, 3 × 3 × 3, S2, P1 & $ 16 \times 4 \times \frac{H}{2} \times \frac{W}{2} $ \\
        UNet Stage 4 conv\_2 & Conv3D, 3 × 3 × 3, S2, P1 & $ 32 \times 2 \times \frac{H}{4} \times \frac{W}{4} $ \\
        UNet Stage 4 conv\_3 & Conv3D, 3 × 3 × 3, S2, P1 & $ 64 \times 1 \times \frac{H}{8} \times \frac{W}{8} $ \\
        
        UNet Stage 4 deconv\_4 & TransposeConv3D, 3 × 3 × 3, S2, P1, OP1 & $ 32 \times 2 \times \frac{H}{4} \times \frac{W}{4} $ \\
        UNet Stage 4 deconv\_5 & TransposeConv3D, 3 × 3 × 3, S2, P1, OP1 & $ 16 \times 4 \times \frac{H}{2} \times \frac{W}{2} $ \\
        UNet Stage 4 deconv\_6 & TransposeConv3D, 3 × 3 × 3, S2, P1, OP1 & $ 8 \times 8 \times H \times W $ \\
        \midrule
    \bottomrule
    \end{tabular}
    }
    \caption{The detailed architecture of the 3D UNet in cost volume regularization. 'S' denotes the stride, 'P' represents the padding and 'OP' is the output\_padding. $\mathbf{V}_{s_{i}} \in \mathbb{R}^{C \times D \times H \times H}$}
    \label{tab:Table.2}
\end{table}

\subsection{Network Architecture of Cost Volume Regularization}
The cost volume regularization is the key part of MVS. According to previous works \cite{b1,b4,b11}, an UNet \cite{b53} structured 3D CNN is applied for cost volume regularization in each stage, which is an hourglass-shaped encoder-decoder architecture. The encoder downsamples cost volume $ \mathbf{V} \in \mathbb{R}^{C \times D \times H \times W}$ into four different scales intermediate feature maps. And decoder upsamples the feature maps back to the original size as an inverse pyramid. For each scale level, the feature map is concatenated with the feature map of the same size in the encoder pyramid along the channel dimension~(transferred by the skip connection). Then it is fed into a transposed 3D CNN with stride 2. The details are shown in Table.\ref{tab:Table.2}.

\subsection{Differentiable Homography Warping}
The differentiable homography warping is the core technique for converting MVS from the traditional domain to the deep learning-based domain. Similar to common practices \cite{b1, b2, b3}, to build feature volumes $ \{ \mathbf{\hat{V}_{i}} \}_{i=1}^{N-1} $, we also utilize the differentiable homography \cite{b1} to warp the feature maps into different fronto-parallel planes of reference camera parameters to construct $ N-1 $ feature volumes $ \{ \mathbf{\hat{V}_{i}} \}_{i=1}^{N-1} $. And we adopt the same cost metric as MVSNet \cite{b1} to aggregate them into a cost volume $ \mathbf{V} $. The equation of differentiable homography is defined as Eq. \eqref{eq:equation_0}:

\begin{equation}
    \label{eq:equation_0}
    \mathbf{H}_{i}^{(d)}=d \mathbf{K}_{i} \mathbf{T}_{i} \mathbf{T}_{r e f}^{-1} \mathbf{K}_{r e f}^{-1}
\end{equation}

\noindent where $ \mathbf{T}, \mathbf{K}$ represent camera extrinsics and intrinsics respectively. $\mathbf{H}_{i}^{(d)}$ represents the homography between the $i^{th}$ the feature map and reference feature map at depth $d$. $\mathbf{T}_{ref}$ represents the extrinsics of reference image.

\section{More Quantitative Results}
\label{sec:B}
In this section, we have provided additional quantitative experiments to prove an adaptive all-fixed depth range can obtain better results with lower consumption for later stages compared to the fixed all-pixel depth range.

Our framework follows a coarse-to-fine strategy for the MVS pipeline (from all-pixel to pixel-wise depth estimation). As we mentioned in introduction, previous methods~(e.g. CasMVSNet, UCSNet) utilize a fixed all-pixel depth range for the depth inference in stage 1, which then influences the computation of the pixel-wise depth range for estimating depth values in subsequent stages. Therefore, the accuracy of the all-pixel depth range directly impacts the quality of the first stage depth map, subsequently affecting the computation of the pixel-wise depth range and the accuracy of the depth map estimation in subsequent stages. To verify this assumption, we conducted a comparative experiment based on UCS-Net. Four groups of all-pixel depth ranges were established: $\textbf{D}_{32}$, $\textbf{D}_{128}$, $\textbf{D}_{512}$, and $\textbf{D}_{adrp}$. The depth interval was set to 2.5, commonly employed in MVSNet, UCS-Net, and so on. Specifically, $\textbf{D}_{32}$ represents an incomplete all-pixel depth range, $\textbf{D}_{128}$ covers most of the scene, $\textbf{D}_{512}$ fully covers the scene, and $\textbf{D}_{adrp}$ denotes the replacement of the all-pixel depth range in the first stage of UCS-Net with the all-pixel depth range predicted by ADRP. For this replacement, depth hypothesis planes were set to $\textbf{D}_{128}$ for the subsequent process. As can be seen from the quantitative results of the Table \ref{tab:Table.R1.2}:

\begin{table}[!h]
    \centering
    \footnotesize
    \setlength{\tabcolsep}{0.8mm}
    \begin{tabular}{c|ccccc}
    \toprule
        Methods & Acc. (mm) $\downarrow$ & Comp. (mm) $\downarrow$ & Overall (mm) $\downarrow$ & Memory (GB) $\downarrow$ & Time (s) $\downarrow$ \\ \hline
        $\mathbf{D}_{32}$ & 0.409 & 0.432 & 0.421 & \textbf{4695MB} & \textbf{0.74s} \\ 
        $\mathbf{D}_{128}$ & 0.341 & 0.347 & 0.344 & 5376MB & 0.83s \\ 
        $\mathbf{D}_{512}$ & \textbf{0.336} & \textbf{0.335} & \textbf{0.336} & 11367MB & 1.71s \\ 
        $\mathbf{D}_{adrp}$ & \underline{0.339} & \underline{0.340} & \underline{0.340} & 5378MB & 0.83s \\
    \bottomrule
    \end{tabular}
    \caption{The quantitative experiment of four different groups of all-pixel depth ranges. The 'underline' represents the sub-optimal results.}
    \label{tab:Table.R1.2}
\end{table}

As shown in Table \ref{tab:Table.R1.2}, we can see that the $\mathbf{D}_{adrp}$ can achieve sub-optimal performance with lower consumption. This demonstrates that a more accurate and reliable all-pixel depth range is superior to a fixed all-pixel depth range. Notably, although $\mathbf{D}_{512}$ yields the best performance in terms of metrics, it performs poorly in terms of efficiency metrics~(Memory:$\mathbf{D}_{512}$-11367MB, Time:$\mathbf{D}_{512}$-1.71s). Hence, it is undesirable to employ a large number of depth hypothesis planes. Additionally, comparing Row 1 and Row 2 reveals that while $\mathbf{D}_{32}$ outperforms $\mathbf{D}_{128}$ in terms of efficiency metrics, it is weaker than $\mathbf{D}_{128}$ in performance metrics~(Acc: $\mathbf{D}_{128}$-0.341 vs $\mathbf{D}_{32}$-0.409, Comp: $\mathbf{D}_{128}$-0.347 vs $\mathbf{D}_{32}$-0.432). Therefore, setting a more suitable all-pixel depth range is more effective. Moreover, comparing Row 2 and Row 4 indicates that a more accurate all-pixel depth range covering the scene can achieve better performance metrics~(Acc: $\mathbf{D}_{128}$-0.341 vs $\mathbf{D}_{adrp}$-0.339, Comp: $\mathbf{D}_{128}$-0.347 vs $\mathbf{D}_{adrp}$-0.340). In summary, a more accurate and reliable all-pixel depth range serves as a stronger foundation for the subsequent pixel-wise depth range, resulting in improved depth estimation outcomes.

\section{More Qualitative Results}
\label{sec:C}
\begin{figure*}[!t]
\centering
\includegraphics[width=1\textwidth]{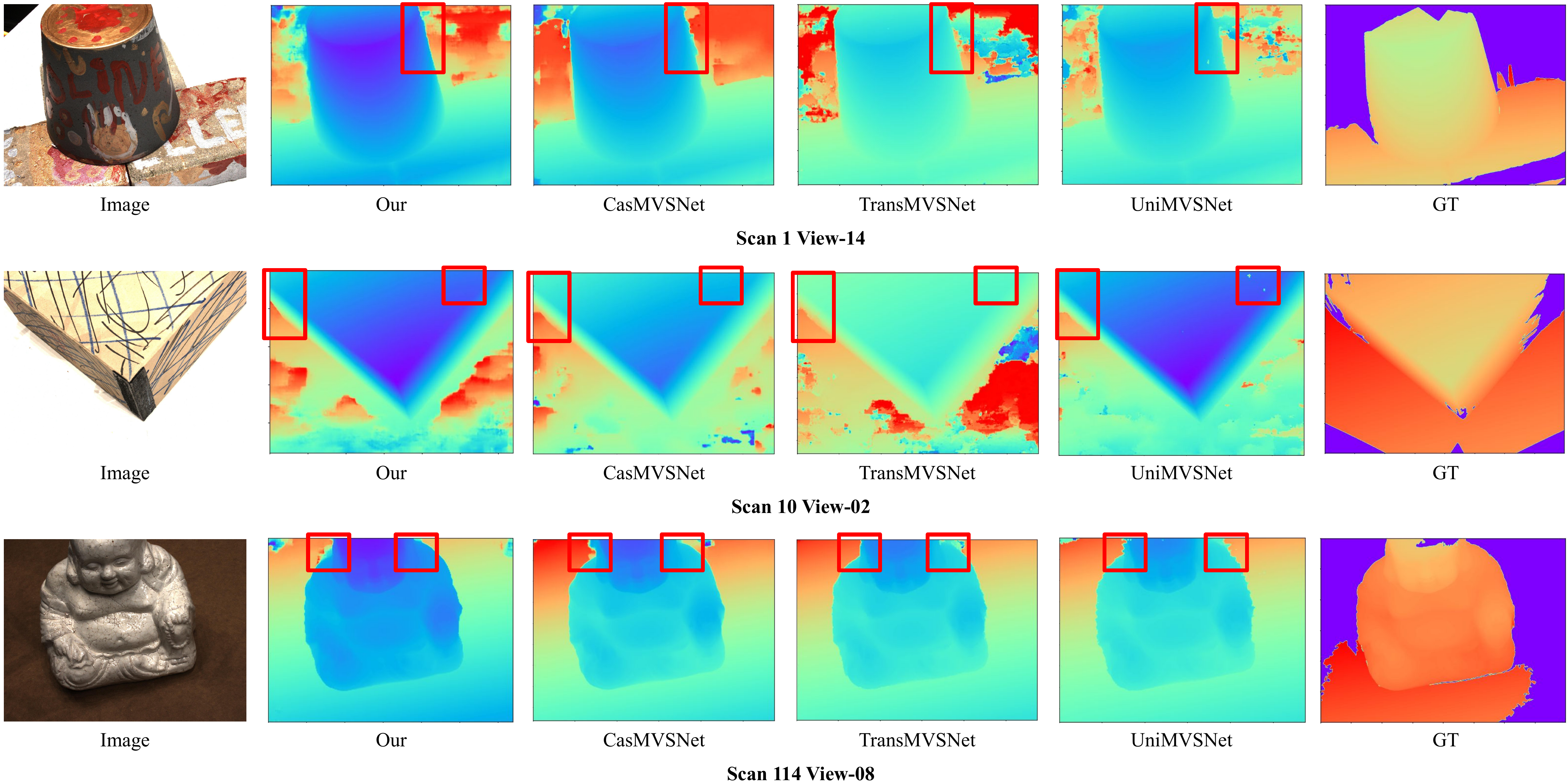}
\caption{Qualitative results of estimated depth on DTU benchmark. The difference in color concentration is mainly due to the maximum and minimum depth values in the depth map~(estimated by different methods). The red boxes are to highlight the important parts.}
\label{fig:qualitative_results}
\end{figure*}

In this section, we exhibit some the qualitative results of depth map on DTU benchmark in Fig.\ref{fig:qualitative_results}. From the qualitative results in the figure, we can see that our model performs well for the reconstruction completeness of each scene.

As shown in the Figure \ref{fig:qualitative_results}, we have selected the depth map of a view from three scans on the DTU benchmark for a qualitative comparison. Upon observation, the depth maps estimated by our method exhibit a similar overall quality to mainstream methods~(e.g. CasMVSNet, TransMVSNet, UniMVSNet). However, our method surpasses these methods in terms of certain details. Specifically, by examining the red boxes in the three scenes, we can see that our method exhibits less depth adhesion between objects and the background compared to other methods~(i.e. depth adhesion refers to the difficulty in distinguishing feature information between foreground and background, which leads to mistaking a part of the background for the foreground and thus estimating the wrong depth value, and vice versa). Among these methods, CasMVSNet has the most serious problem of depth adhesion, while UniMVSNet has some jagged depth distribution in the depth estimation of object edges, and occurs some noisy depth values on the surface of the object.

\section{More Reconstruction Results}
\label{sec:D}
\begin{figure*}[!ht]
\centering
\includegraphics[width=1\textwidth]{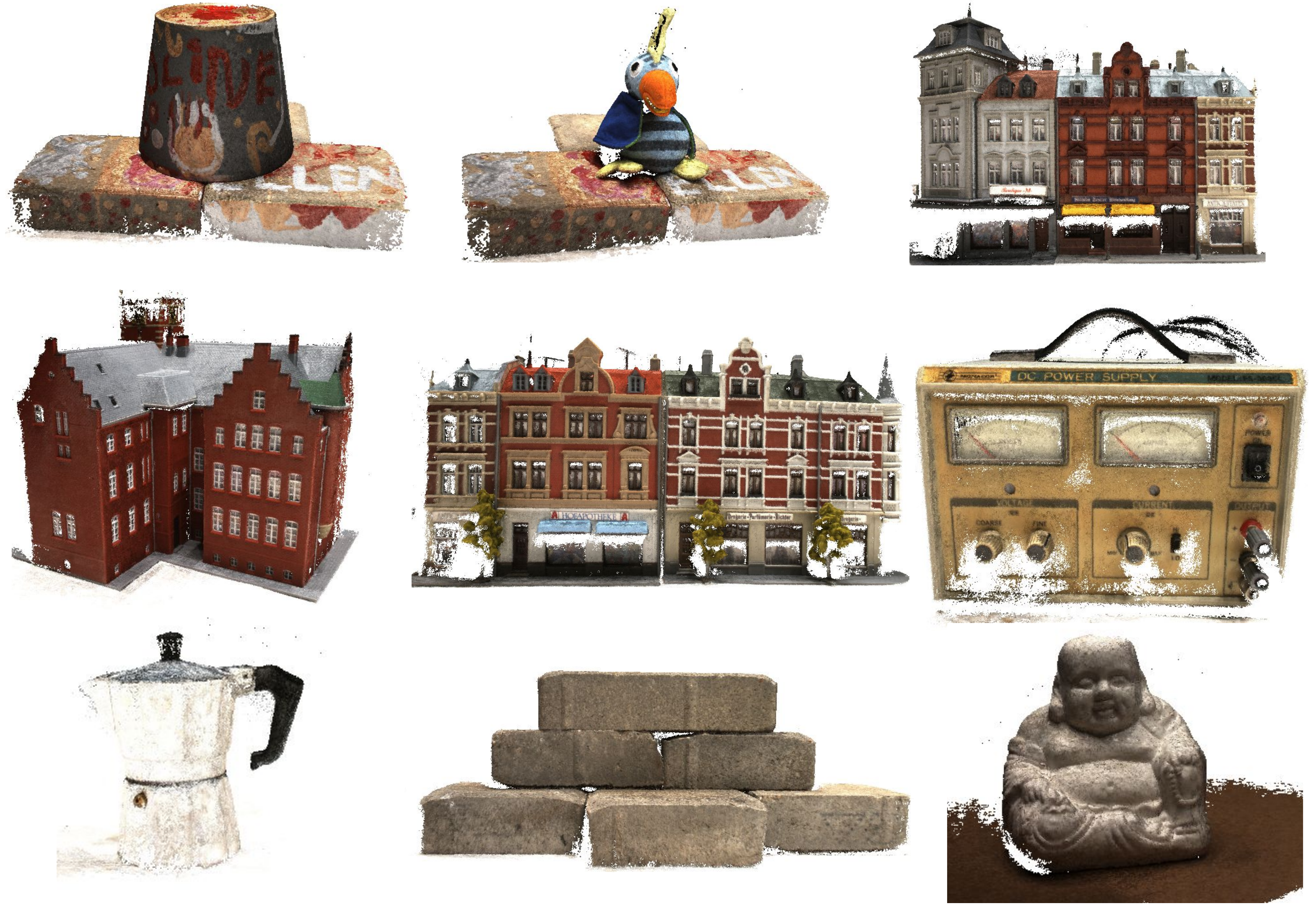}
\caption{More qualitative results on DTU benchmark.}
\label{fig:Figure 4}
\end{figure*}

\begin{figure*}[!ht]
\centering
\includegraphics[width=1\textwidth]{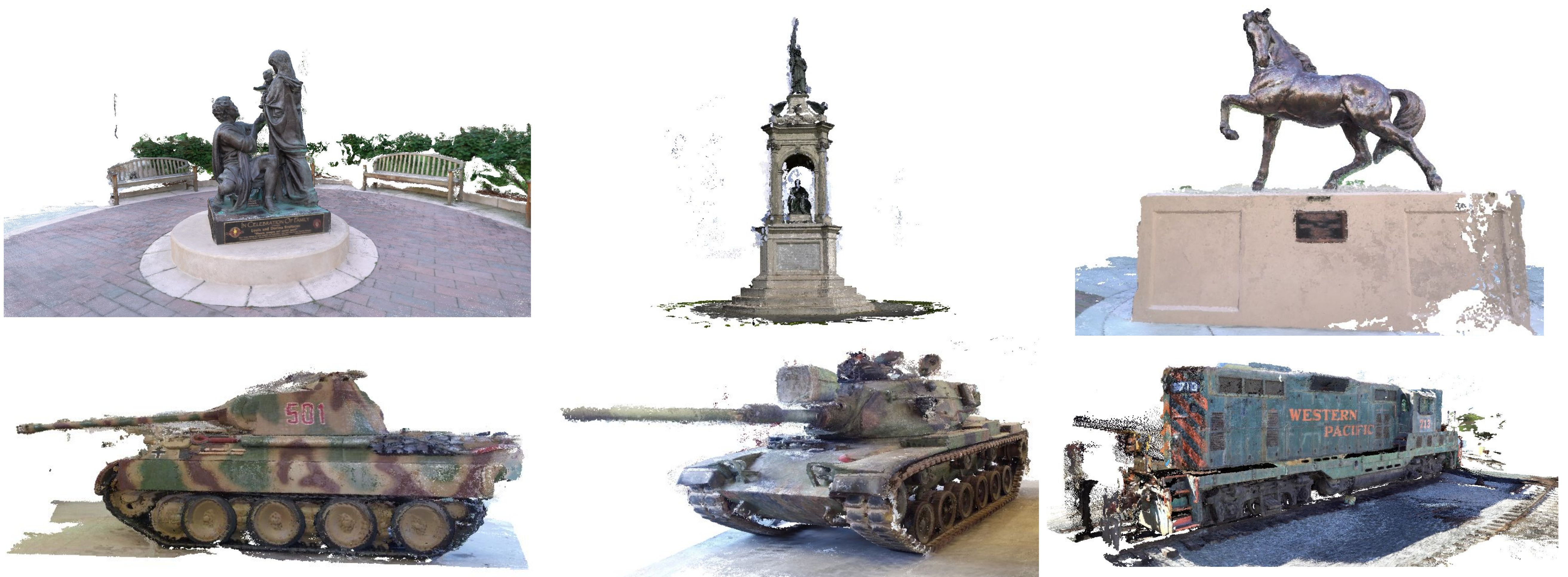}
\caption{More qualitative results on Tanks and Temples benchmark.}
\label{fig:Figure 5}
\end{figure*}

In this section, we show the point cloud results on DTU benchmark in Fig.\ref{fig:Figure 4}. From the reconstruction results in the figure, we can see that our model performs well for the reconstruction completeness of each scene. Some results of Tanks and Temples benchmark are shown in Fig.\ref{fig:Figure 5}.

\clearpage

\bibliographystyle{elsarticle-num}
\bibliography{mybibfile}